\documentclass[a4paper,fleqn]{cas-dc}

\usepackage[numbers]{natbib}
\usepackage{amsmath,amssymb,amsfonts}
\usepackage{algorithm}
\usepackage{algorithmic}
\usepackage{array}
\usepackage{subfig}
\usepackage{textcomp}
\usepackage{booktabs}
\usepackage{makecell}
\usepackage{multirow}
\usepackage{csquotes}
\usepackage{stfloats}
\usepackage{utfsym}
\usepackage{url}
\usepackage{verbatim}
\usepackage{graphicx}
\usepackage{threeparttable}
\usepackage{xcolor}
\usepackage{tikz}
\usepackage{pifont}
\usepackage{rotating}
\usepackage{placeins}

\newcommand{\best}[1]{\textcolor{blue!95!black}{\textbf{#1}}}
\newcommand{\second}[1]{#1}

\newcommand{\std}[1]{\ensuremath{{}\pm\!{\scriptstyle #1}}}
\newcommand{\sstd}[1]{{\tiny\ensuremath{\pm #1}}}
\def\BibTeX{{\rm B\kern-.05em{\sc i\kern-.025em b}\kern-.08em T\kern-.1667em\lower.7ex\hbox{E}\kern-.125emX}}

\begin{document}
\let\WriteBookmarks\relax
\def\floatpagepagefraction{1}
\def\textpagefraction{.001}

\shorttitle{Registration-Grounded Spectral Fusion}
\shortauthors{P. Jie et~al.}

\title[mode=title]{Registration-Grounded Spectral Fusion for Unregistered WLI/NBI Endoscopic Lesion Segmentation}
\tnotemark[1]

\author[1]{Pengyu Jie}

\author[1]{Wanquan Liu}
\cormark[1]
\ead{liuwq63@mail.sysu.edu.cn}

\author[2]{Rui He}
\author[3]{Pengcheng Li}
\author[2]{Weiping Wen}
\author[4]{Deyu Meng}
\author[5]{Junwei Han}

\author[1]{Chenqiang Gao}
\cormark[1]
\ead{gaochq6@mail.sysu.edu.cn}

\affiliation[1]{
  organization={School of Intelligent Systems Engineering, Shenzhen Campus of Sun Yat-sen University},
  city={Shenzhen},
  postcode={518107},
  state={Guangdong},
  country={P.R. China}
}

\affiliation[2]{
  organization={Department of Otolaryngology, The First Affiliated Hospital of Sun Yat-sen University},
  city={Guangzhou},
  postcode={510000},
  country={P.R. China}
}

\affiliation[3]{
  organization={School of Computer Science and Technology, Hainan University},
  city={Haikou},
  postcode={570228},
  country={China}
}

\affiliation[4]{
  organization={School of Mathematics and Statistics, Xi'an Jiaotong University},
  city={Xi'an},
  postcode={710049},
  country={China}
}
 
\affiliation[5]{
  organization={Chongqing University of Posts and Telecommunications},
  city={Chongqing},
  postcode={400065},
  country={China}
}

\cortext[cor1]{Corresponding authors.}

\tnotetext[1]{This work is supported in part by the Shenzhen Fundamental Research Program (Grant No. JCYJ20240813151216022).}
\begin{abstract}
White-light imaging (WLI) and narrow-band imaging (NBI) provide complementary cues for endoscopic lesion segmentation, yet their sequential acquisition often results in spatially inconsistent observations. The central challenge is not merely to establish cross-modal correspondence, but to determine how complementary information should be incorporated when such correspondence is uncertain. To address this issue, we propose a correspondence-conditioned, modality-asymmetric complex spectral fusion framework for paired-but-unregistered WLI/NBI segmentation. The framework first establishes bidirectional feature correspondence through topology-regularized registration and derives local reliability from post-warping feature discrepancies. Rather than simply reweighting fused features, the estimated reliability directly controls the construction of modality-asymmetric polar representations, where WLI-derived cues primarily parameterize appearance-related moduli and NBI-derived cues parameterize structure-sensitive angular components. The resulting complex representations are further refined through Fourier-domain recalibration with reliability-biased residual reconstruction. Experiments on FAHSYSU and PICCOLO demonstrate competitive improvements over existing segmentation methods. Module ablations, modality-role reversal, and reliability analyses further support the effectiveness of correspondence-conditioned polar representation learning.
\end{abstract}


\maketitle
\section{Introduction}\label{sec:introduction}
Endoscopic examination plays an important role in the screening and assessment of mucosal lesions, and accurate lesion segmentation provides essential support for early detection and diagnosis \cite{Tiwari2025,jiepy_eClinicalMedicine}. In routine examinations, clinicians often inspect suspicious lesions using both white-light imaging (WLI) and narrow-band imaging (NBI). WLI provides color appearance and contextual morphology for region-level lesion localization, whereas NBI enhances mucosal and microvascular patterns, making boundary-related structures more distinguishable \cite{jiepy_eClinicalMedicine}. However, WLI/NBI observations of the same lesion are typically paired but unregistered due to sequential hand-held acquisition, viewpoint changes, and nonrigid tissue deformation, as illustrated in Fig.~\ref{fig:introduction_figure}(a) and (b). The lack of task-relevant pixel-to-pixel correspondence makes direct WLI/NBI fusion prone to misleading cross-modal interference. This difficulty highlights a gap between clinical dual-modality observation and current automatic segmentation practice. While paired-but-unregistered WLI/NBI fusion has recently received some attention, it remains much less explored than single-modality endoscopic lesion segmentation \cite{jiepy_TCSVT,MCSNet,jiepy_eClinicalMedicine,endo_tmi_1,endo_tmi_2}.

\begin{center}
    \centering
    \includegraphics[width=\columnwidth,height=0.36\textheight,keepaspectratio]{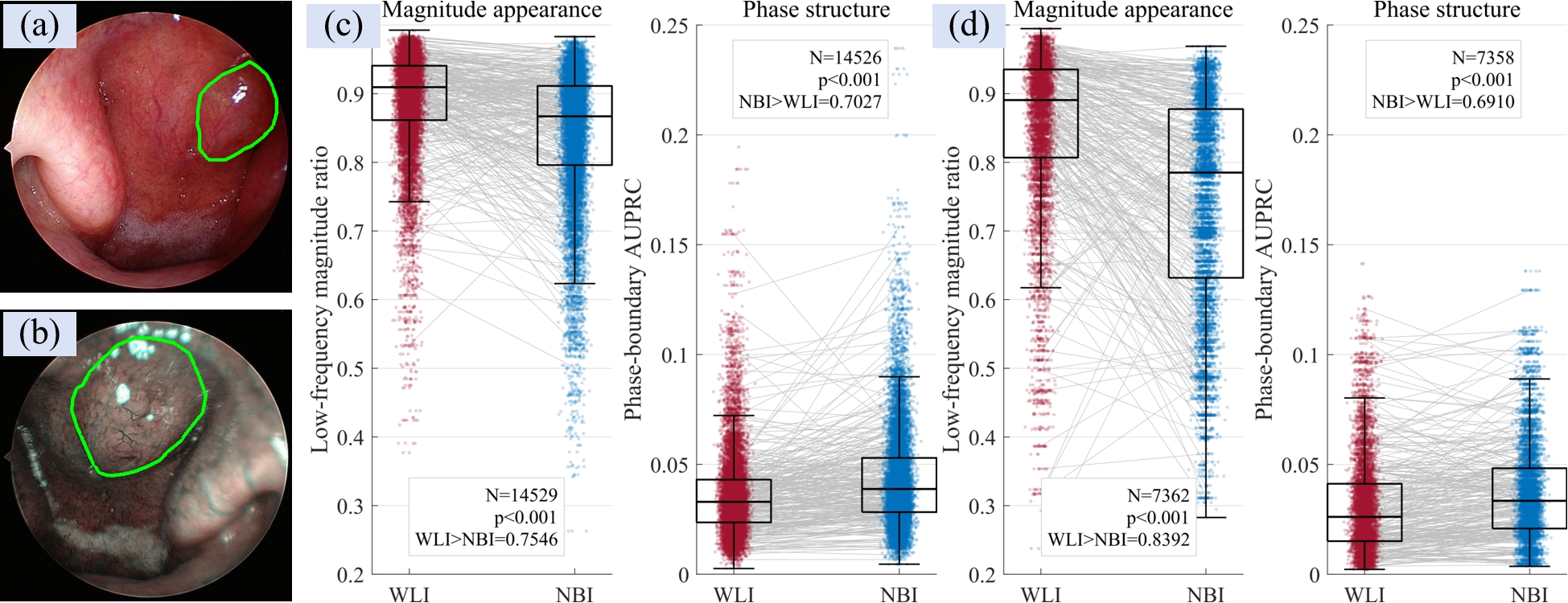}
    \captionof{figure}{Illustration of paired-but-unregistered WLI/NBI images and modality-role analysis. 
    (a,b) Representative WLI and NBI images of the same lesion show noticeable spatial mismatch caused by hand-held acquisition. 
    (c,d) Dataset-level analysis on FAHSYSU (Paired) and PICCOLO (Paired) shows that WLI has a higher low-frequency magnitude ratio than NBI, while NBI-derived phase responses show stronger boundary correspondence than WLI-derived responses. 
    These image-level spectral observations motivate a modality-asymmetric inductive bias for constructing learned spatial complex representations, rather than directly exchanging Fourier magnitude and phase components.}
    \label{fig:introduction_figure}
\end{center}

Although dual modality WLI/NBI segmentation is clinically motivated, effectively exploiting paired-but unregistered observations remains challenging due to cross-modal spatial mismatch and heterogeneous imaging cues. Under unregistered acquisition, naive concatenation, position-wise attention, or direct feature fusion may mix non-corresponding regions, leading to boundary drift or negative transfer \cite{Su_2025_ICCV,xu2023murf,huang2022reconet,chen2024weakly}. 
Misalignment-aware multimodal segmentation has been investigated through flow-guided registration with frequency-aware interaction in BRENet \cite{BRENet2025}, anatomical-position-guided patch association in Patch2Space \cite{Patch2Space2026}, and affine correction with cross-modal attention in ACMSegNet \cite{ACMSegNet2025}.
In the endoscopic domain, ADFNet is a representative attempt for unregistered WLI/NBI segmentation, which enhances cross-modal semantic compatibility by separating shared features from modality-specific cues \cite{ADFNet}. Beyond endoscopy, methods alleviate cross-modal discrepancy through generative modeling, feature alignment, or robust fusion strategies \cite{UMSCS,Su_2025_ICCV,DONG2025113427,wang2025rfsc}. 
Nevertheless, paired-but-unregistered WLI/NBI fusion must address two coupled issues. First, spatial alignment alone does not determine how strongly cross-modal information should be incorporated at locations with uncertain correspondence \cite{ADFNet,UMSCS,DONG2025113427,wang2025rfsc}.
Consequently, residual local mismatch may cause unreliable feature fusion around lesion boundaries and fine mucosal structures. 
Second, the complementary roles of WLI and NBI motivate assigning their cues to distinct representation components, rather than relying on undifferentiated feature aggregation.
In paired-but-unregistered endoscopy, these issues are coupled: unreliable spatial correspondence can amplify cross-modal interference, while undifferentiated aggregation can obscure modality-specific appearance and structural cues. Therefore, effective WLI/NBI segmentation requires reliability-aware correspondence modeling and role-aware complementary fusion.

To address these challenges, we propose a correspondence-conditioned complex spectral fusion framework for paired-but-unregistered WLI/NBI lesion segmentation. Our central insight is that geometric alignment alone is insufficient for effective cross-modal fusion, as the reliability of local correspondence should also determine how complementary modality cues are incorporated into feature representations. This design is motivated by the dataset-level spectral observations in Fig.~\ref{fig:introduction_figure}(c) and (d), where WLI exhibits a higher low-frequency Fourier-magnitude ratio than NBI, whereas NBI-derived phase responses show stronger correspondence with lesion boundaries. Rather than assuming a direct equivalence between image-level Fourier components and learned complex features, we leverage these observations as an inductive bias for modality-asymmetric polar representation learning. Specifically, we develop Reliability-Aware Diffeomorphic Registration Grounding (R-DRG) to establish topology-regularized cross-modal feature correspondence and estimate local reliability from post-warping feature discrepancies. Building upon this correspondence-aware geometric basis, we introduce Reliability-Guided Complex Spectral Fusion (R-CSF), which assigns WLI-derived cues primarily to the modulus and NBI-derived cues to the angular components of learned spatial complex representations. Unlike conventional reliability-weighted feature aggregation, the estimated correspondence reliability directly conditions the contribution of aligned cross-modal cues to individual polar components, allowing modality-specific information to be preserved where correspondence is uncertain. The resulting complex representations undergo frequency-selective Fourier-domain recalibration, followed by reliability-biased residual reconstruction. In this way, registration establishes candidate correspondences, modality-role assignment defines how complementary cues are represented, and reliability governs their incorporation during both polar construction and subsequent spectral refinement.
The main contributions of this work are summarized as follows:
\begin{itemize}
    \item We introduce a correspondence-conditioned, modality-asymmetric polar representation learning formulation for paired-but-unregistered WLI/NBI segmentation, in which local correspondence reliability governs the contribution of complementary modality cues to distinct polar components rather than merely weighting aggregated features.

    \item We instantiate this formulation through Reliability-Aware Diffeomorphic Registration Grounding (R-DRG) and Reliability-Guided Complex Spectral Fusion (R-CSF). R-DRG establishes topology-regularized feature correspondence and estimates local reliability, while R-CSF uses this reliability to condition modality-asymmetric polar construction and residual reconstruction after Fourier-domain recalibration.

    \item Extensive experiments on paired WLI/NBI endoscopic datasets demonstrate improved segmentation performance over representative baselines. Module ablations, registration comparisons, modality-role reversal, and reliability analyses further support the effectiveness of the proposed correspondence-conditioned fusion design.
\end{itemize}

\section{Related Work}
\subsection{WLI/NBI Endoscopic Lesion Segmentation}
WLI and NBI have been widely investigated in endoscopic lesion analysis, including lesion detection, diagnosis, classification, segmentation, and modality-related enhancement or synthesis tasks \cite{jiepy_eClinicalMedicine,hu2025holistic,lei2025virtual,chen2025evaluation}. These studies confirm the clinical relevance of both imaging modes, but most of them either focus on a single modality or use the other modality as auxiliary diagnostic information, rather than performing explicit dual-modality fusion for lesion segmentation.
Only a few studies have directly explored WLI/NBI fusion for endoscopic lesion segmentation. MCSNet introduces lesion-oriented modeling to improve endoscopic lesion delineation \cite{MCSNet}, while ADFNet further investigates multimodal WLI/NBI representation learning through progressive disentanglement, distribution alignment, and contrastive learning \cite{ADFNet}. However, existing WLI/NBI fusion methods mainly focus on semantic complementarity or feature compatibility, and insufficiently address the paired-but-unregistered nature of hand-held endoscopic acquisition. In particular, dense and topology-preserving cross-modal correspondence is rarely established before fusion, and modality-specific complementary roles are not explicitly modeled. This motivates a framework that jointly considers geometric mismatch and cross-modal complementary fusion for WLI/NBI lesion segmentation.

\subsection{Deformable Registration under Misalignment}
Deformable registration has been widely studied for establishing dense spatial correspondence in medical images. Recent learning-based registration methods increasingly emphasize smooth, invertible, and topology-preserving transformations, especially through diffeomorphic modeling, to improve deformation regularity and anatomical plausibility \cite{chen2024mirsurvey,joshi2023r2net,sun2024nir,chen2025spatialreg,fogarollo2025implicit}. In addition, joint registration-segmentation frameworks show that deformation estimation can serve as a task-driven intermediate mechanism, where segmentation-related constraints guide the learned correspondence and benefit downstream delineation \cite{khor2023anatomically,DCRS,Fan_2024_CVPR}.
Despite these advances, WLI/NBI endoscopic lesion segmentation differs from conventional intra-modality or anatomically standardized registration tasks. The two modalities are only paired and their mismatch is jointly caused by cross-modal appearance discrepancy, asynchronous hand-held acquisition, viewpoint variation, illumination changes, and soft-tissue deformation. Under such conditions, unconstrained offsets or attention-only alignment may introduce unstable correspondences, especially around lesion boundaries. Therefore, feature-level diffeomorphic registration is more suitable for this task, not for recovering a physically exact inter-frame motion field, but for providing a smooth and topology-preserving geometric basis before cross-modal fusion.

\begin{figure*}[t]
\vspace{-2mm}
	\centering 
	\includegraphics[width=0.8\textwidth]{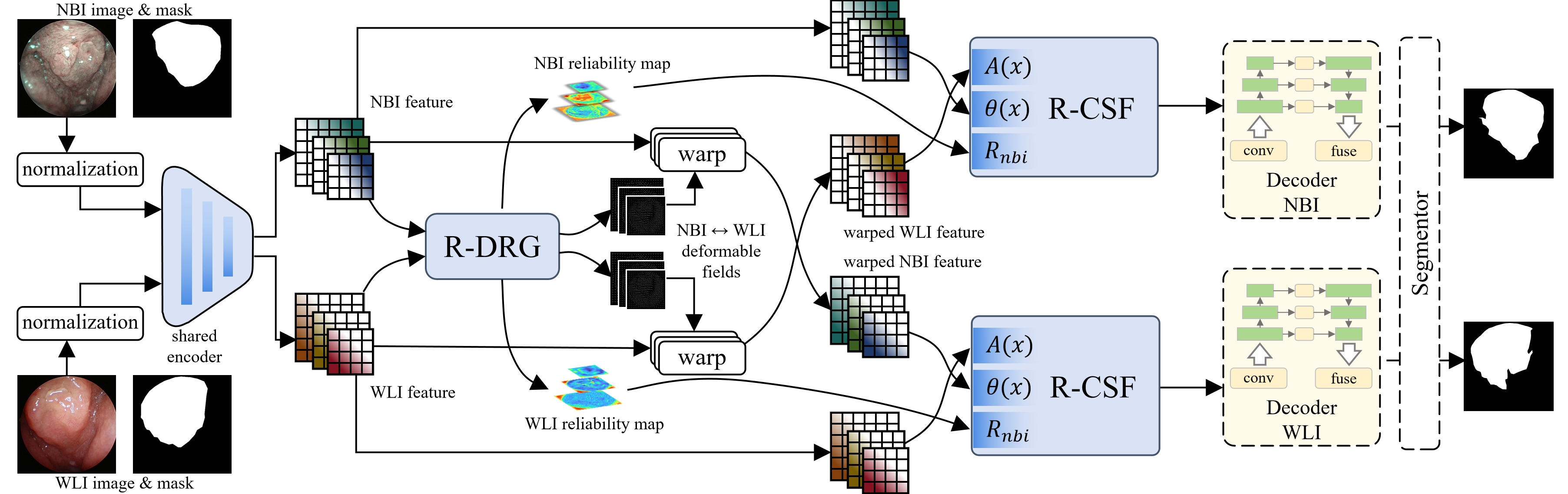}
	\caption{Overview of the proposed framework. The framework consists of a modality-aware shared encoder, Reliability-Aware Diffeomorphic Registration Grounding (R-DRG), Reliability-Guided Complex Spectral Fusion (R-CSF), and dual-branch segmentation decoders. R-DRG establishes topology-preserving cross-modal feature correspondence and derives correspondence reliability maps, while R-CSF uses these maps to guide role-aware complex spectral fusion of WLI-dominant appearance cues and NBI-dominant structural cues.}

	\label{fig:overall_framework}
\vspace{-5mm}
\end{figure*}

\subsection{Cross-Modal Fusion}
Multimodal fusion methods commonly integrate information through input-level concatenation, weighted aggregation, attention-based interaction, or multi-scale feature exchange \cite{CAO2025114597,JIANG2026112723,zhang2025mgnet}. These strategies are effective in many multimodal scenarios, but they often rely on task-relevant geometric guidance. When applied to paired but unregistered images, direct spatial-domain fusion may mix non-corresponding regions, leading to boundary drift, artifact amplification, or negative transfer \cite{Su_2025_ICCV,xu2023murf,huang2022reconet,chen2024weakly}. Recent misalignment-aware fusion and registration-fusion methods attempt to improve robustness by introducing correspondence modeling or progressive alignment \cite{DONG2025113427,CHEN2026103960,wang2025rfsc}. Nevertheless, their subsequent interaction is mainly performed in the spatial or real-valued feature domain.
For WLI/NBI endoscopy, fusion should consider the heterogeneous roles of the two modalities. Frequency-domain learning offers a natural way to distinguish global appearance and local structural cues, and studies have explored frequency-aware fusion or magnitude-phase learning for dense prediction \cite{10648934,yan2024decoupling,liu2025frequency}. However, explicit frequency-aware and complex-valued fusion remains underexplored for paired but unregistered WLI/NBI lesion segmentation. 
This motivates our registration-grounded complex spectral fusion, which uses the observed WLI/NBI modality asymmetry as an inductive bias for constructing learned spatial complex representations, followed by explicit Fourier-domain recalibration.

\section{Method}\label{sec_framework}
\subsection{Problem Formulation and Framework Overview}

Given a paired WLI--NBI sample of the same lesion, denoted as $(x_{wli},x_{nbi},y_{wli},y_{nbi})$, where $x_{wli}$ and $x_{nbi}$ are the WLI and NBI images and $y_{wli},y_{nbi}\in\{0,1\}^{H\times W}$ are the corresponding binary lesion masks, our goal is to learn a dual-branch segmentation mapping:
\begin{equation}
\resizebox{0.55\linewidth}{!}{$
\begin{aligned}
f_{\Theta}:(x_{wli},x_{nbi})\mapsto(\hat{y}_{wli},\hat{y}_{nbi}),
\end{aligned}
$}
\end{equation}
where $\hat{y}_{wli},\hat{y}_{nbi}\in[0,1]^{H\times W}$ are the predicted lesion probability maps for the two modalities.
Different from conventional aligned multimodal segmentation, the WLI--NBI pairs considered in this work are paired but not pixel-wise registered, since the two images depict the same lesion but are acquired under different imaging modalities and viewpoints.
This setting introduces both cross-modal appearance discrepancy and geometric mismatch, making it essential to exploit WLI/NBI complementarity while suppressing unreliable interaction between locally non-corresponding regions.

As shown in Fig.~\ref{fig:overall_framework}, the proposed framework consists of modality-aware pyramid encoding, Reliability-Aware Diffeomorphic Registration Grounding (R-DRG), Reliability-Guided Complex Spectral Fusion (R-CSF), and dual-branch segmentation decoding.
For each modality $m\in\{wli,nbi\}$, a shared pyramid encoder with lightweight modality-specific adaptation extracts multi-scale features as:
\begin{equation}
\resizebox{0.55\linewidth}{!}{$
\begin{aligned}
\{f_m^{l}\}_{l=1}^{L}=E(\gamma_m\odot x_m+\beta_m;\eta_m),
\end{aligned}
$}
\end{equation}
where $\gamma_m$ and $\beta_m$ are learnable channel-wise affine parameters and $\eta_m$ denotes the modality-specific adapter parameters.
This design maintains a common representation space for cross-modal interaction while preserving modality-specific characteristics.
The encoded features are first processed by R-DRG to establish topology-regularized cross-modal correspondence, which provides aligned representations for subsequent spectral fusion. In our implementation, the deformation field is estimated at the deepest pyramid level (16 $\times$ 16) and then resized to other feature levels for multi-scale feature warping. 
Based on the warped features and reliability maps, R-CSF constructs modality-asymmetric spatial complex representations through reliability-guided polar parameterization, followed by frequency-selective Fourier-domain recalibration to integrate complementary WLI/NBI cues.
Finally, the enhanced features are decoded by two modality-specific decoders to predict $\hat{y}_{wli}$ and $\hat{y}_{nbi}$, and the whole framework is optimized end-to-end with segmentation and alignment-related objectives.

\subsection{Reliability-Aware Diffeomorphic Registration Grounding, R-DRG}

\begin{figure}
    \centering  
    \includegraphics[width=\columnwidth,height=0.36\textheight,keepaspectratio]{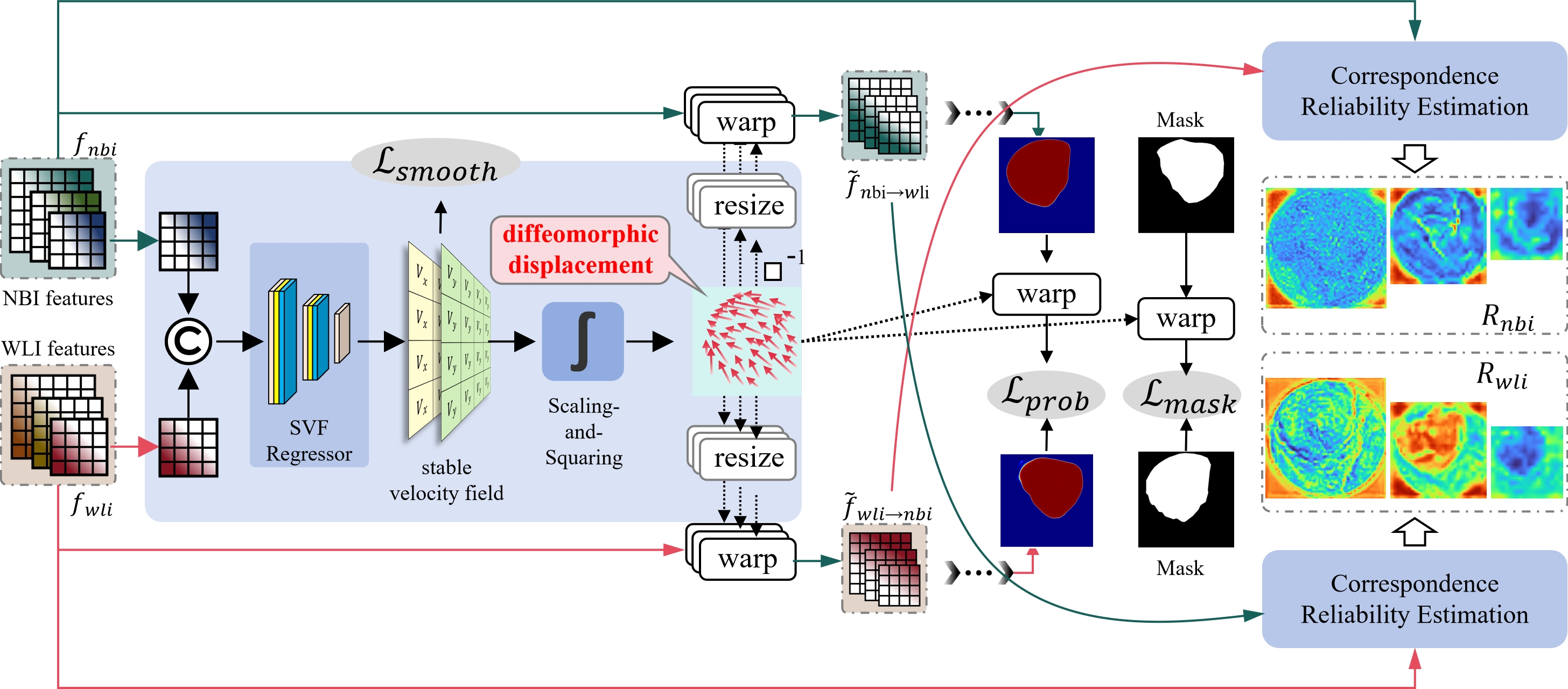}
    \caption{Illustration of the proposed Reliability-Aware Diffeomorphic Registration Grounding (R-DRG) module.}
    \label{fig:RDRG}
\vspace{-5mm}
\end{figure}

Although each WLI--NBI pair depicts the same lesion, the two images are not spatially aligned due to viewpoint variation, tissue deformation, and handheld acquisition.
Direct fusion under such paired but unregistered conditions may mix non-corresponding regions, especially around lesion boundaries and fine mucosal structures.
We therefore propose a Reliability-Aware Diffeomorphic Registration Grounding (R-DRG) module to establish topology-regularized feature correspondence and estimate where cross-modal interaction is reliable before spectral fusion, as illustrated in Fig.~\ref{fig:RDRG}.
 
Rather than registering raw images, R-DRG performs registration in feature space, where semantic representations are more robust to low-level WLI/NBI appearance discrepancy.
Given the encoded feature pyramids $\{f_{wli}^{l}\}_{l=1}^{L}$ and $\{f_{nbi}^{l}\}_{l=1}^{L}$, a stationary velocity field is estimated from the deepest feature level and integrated with opposite signs to obtain bidirectional diffeomorphic transformations:
\begin{equation}
\resizebox{0.35\linewidth}{!}{$
\begin{aligned}
&v^{L} = \mathcal{R}\big([f_{wli}^{L},f_{nbi}^{L}]\big), \\
&\phi_{nbi\rightarrow wli}^{L} = \exp(v^{L}), \\
&\phi_{wli\rightarrow nbi}^{L} = \exp(-v^{L}).
\end{aligned}
$}
\end{equation}
Here, $[\cdot,\cdot]$ denotes channel-wise concatenation, $\mathcal{R}(\cdot)$ is the velocity-field regressor, and $\exp(\cdot)$ is implemented by scaling-and-squaring integration. $\mathcal(\cdot)$ is a convolutional stationary velocity field (SVF) regressor comprising four $3\times3$ convolutional blocks with GroupNorm and GELU activation, followed by a $3\times3$ output convolution. The channel dimensions are $512\!{\scriptstyle\to}\!128\!{\scriptstyle\to}\!128\!{\scriptstyle\to}\!128\!{\scriptstyle\to}\!128\!{\scriptstyle\to}\!2$, with all convolutions using a stride of 1, padding of 1, and dilation of 1, yielding a nominal convolutional receptive field of $11\times11$ feature-map locations. The output layer is zero-initialized to stabilize deformation learning at the early training stage. No spatial downsampling is applied in this module to preserve dense correspondence estimation.
This single-SVF parameterization encourages smooth, invertible, and topology-preserving cross-modal correspondence while avoiding the need to estimate two independent deformation fields.

The deformation fields are rescaled to each pyramid resolution, and the multi-scale features are warped bidirectionally as:
\begin{equation}
\resizebox{0.45\linewidth}{!}{$
\begin{aligned}
\tilde{f}_{nbi\rightarrow wli}^{\,l}
&= \mathcal{W}\!\left(f_{nbi}^{l},\phi_{nbi\rightarrow wli}^{\,l}\right), \\
\tilde{f}_{wli\rightarrow nbi}^{\,l}
&= \mathcal{W}\!\left(f_{wli}^{l},\phi_{wli\rightarrow nbi}^{\,l}\right).
\end{aligned}
$}
\end{equation}
Here, $\mathcal{W}(\cdot,\cdot)$ denotes differentiable feature warping, and the warped features provide geometry-aware cross-modal candidates for the subsequent fusion stage.

However, smooth warping does not guarantee that all regions are equally reliable, since WLI/NBI pairs may still contain local non-correspondence caused by viewpoint changes, field-of-view differences, illumination variations, or modality-specific appearance.
To avoid blindly injecting unreliable cross-modal cues, R-DRG further computes correspondence reliability maps from the projected feature discrepancy after warping:
\begin{equation}
\resizebox{0.7\linewidth}{!}{$
\begin{aligned}
R_{wli}^{l}
&=
R_{\mathrm{valid},wli}^{l}
\cdot
\exp\!\left(
-\frac{
\left\|
p^{l}(f_{wli}^{l}) -
p^{l}(\tilde{f}_{nbi\rightarrow wli}^{\,l})
\right\|_{1}
}{\tau_f}
\right), \\
R_{nbi}^{l}
&=
R_{\mathrm{valid},nbi}^{l}
\cdot
\exp\!\left(
-\frac{
\left\|
p^{l}(f_{nbi}^{l}) -
p^{l}(\tilde{f}_{wli\rightarrow nbi}^{\,l})
\right\|_{1}
}{\tau_f}
\right).
\end{aligned}
$}
\end{equation}
Here, $p^{l}(\cdot)$ is a correspondence reliability estimation module that maps WLI and NBI features into a shared comparison space, and $\tau_f$ is a temperature parameter. 
$R_{\mathrm{valid},a\rightarrow b}^{l}$ is a non-learnable validity mask derived from the sampling grid of $\phi_{a\rightarrow b}^{l}$, indicating whether each target-grid location samples from a valid source feature region after warping.
The reliability maps are not used as additional segmentation supervision, but serve as correspondence priors for the following complex spectral fusion module.
Regions with high reliability are allowed to receive stronger cross-modal spectral cues, whereas regions with low reliability are encouraged to rely more on modality-specific representations.
In this way, R-DRG not only establishes registration-grounded feature correspondence, but also identifies where cross-modal spectral interaction is trustworthy, thereby reducing negative transfer caused by locally mismatched WLI/NBI observations.

\subsection{Reliability-Guided Complex Spectral Fusion, R-CSF}

Given the warped features and reliability maps produced by R-DRG, we further propose Reliability-Guided Complex Spectral Fusion (R-CSF) to exploit WLI/NBI complementarity through a learned complex representation followed by frequency-selective recalibration. Instead of directly concatenating real-valued features or exchanging Fourier coefficients, R-CSF constructs modality-role-aware polar representations in the spatial feature domain, where WLI-derived cues primarily parameterize appearance-related moduli and NBI-derived cues primarily parameterize structure-sensitive angular components. This asymmetric assignment is motivated by the image-level spectral observations in Fig.~\ref{fig:introduction_figure}, rather than an assumed equivalence between spatial polar components and Fourier magnitude–phase spectra. Correspondence reliability further determines the contribution of aligned cross-modal cues during polar construction, allowing the model to retain modality-specific information where correspondence is uncertain.

For the WLI branch, the WLI feature provides the appearance-dominant magnitude, while the phase response is adaptively interpolated between the warped NBI structural cue and the WLI self cue according to the reliability map $R_{wli}^{l}$.
For the NBI branch, the NBI feature provides the structure-sensitive phase, while the magnitude response is adaptively interpolated between the warped WLI appearance cue and the NBI self cue according to $R_{nbi}^{l}$:
\begin{equation}\label{eq:role_assignment}
\resizebox{0.85\linewidth}{!}{$
\begin{aligned}
A_{wli}^{l} &= \mathrm{softplus}\!\left(h_{A}^{w}(f_{wli}^{l})\right), \\
\theta_{wli}^{l} &= R_{wli}^{l}\odot \pi\tanh\!\left(h_{\theta}^{c}(\tilde{f}_{nbi\rightarrow wli}^{\,l})\right)
+ (1-R_{wli}^{l})\odot \pi\tanh\!\left(h_{\theta}^{s}(f_{wli}^{l})\right), \\
A_{nbi}^{l} &= R_{nbi}^{l}\odot \mathrm{softplus}\!\left(h_{A}^{c}(\tilde{f}_{wli\rightarrow nbi}^{\,l})\right)
+ (1-R_{nbi}^{l})\odot \mathrm{softplus}\!\left(h_{A}^{s}(f_{nbi}^{l})\right), \\
\theta_{nbi}^{l} &= \pi\tanh\!\left(h_{\theta}^{n}(f_{nbi}^{l})\right).
\end{aligned}
$}
\end{equation}
Here, $h_{A}^{w}(\cdot)$, $h_{A}^{c}(\cdot)$, $h_{A}^{s}(\cdot)$, $h_{\theta}^{c}(\cdot)$, $h_{\theta}^{s}(\cdot)$, and $h_{\theta}^{n}(\cdot)$ are lightweight learnable projections.
The reliability maps are broadcast along the channel dimension.
This design preserves the non-exchangeable WLI/NBI roles while allowing the model to fall back to modality-specific cues when cross-modal correspondence is unreliable.

The branch-specific complex features are then constructed as:
\begin{equation}\label{eq:complex_feature}
\resizebox{0.55\linewidth}{!}{$
\begin{aligned}
z_{wli}^{l} &= A_{wli}^{l}\left(\cos\theta_{wli}^{l}+i\sin\theta_{wli}^{l}\right), \\
z_{nbi}^{l} &= A_{nbi}^{l}\left(\cos\theta_{nbi}^{l}+i\sin\theta_{nbi}^{l}\right).
\end{aligned}
$}
\end{equation}
The softplus operation enforces non-negative moduli, while the bounded tanh mapping stabilizes the learned angular responses. Importantly, the modulus and angle in Eq.~\ref{eq:complex_feature} describe the polar coordinates of the constructed complex features at each spatial location, rather than the magnitude and phase of their Fourier transforms. Thus, the image-level spectral observations in Fig. 1 serve as a modality-role prior without imposing a direct correspondence between the learned polar components and Fourier coefficients. The resulting spatial complex features are subsequently transformed into the frequency domain for frequency-selective recalibration, as described below.

For each branch $m\in\{wli,nbi\}$, we transform the complex feature into the frequency domain and perform frequency-selective complex recalibration:
\begin{equation}
\resizebox{0.85\linewidth}{!}{$
\begin{aligned}
&Z_{m}^{l} = \mathcal{F}(z_{m}^{l}), \\
&G_{m}^{l}(\xi,\zeta) = \sum_{k=1}^{K} M_{k}(\xi,\zeta)c_{m,k}^{l}, \\
&s_{m}^{l} = \mathrm{GAP}\!\left([|Z_{m}^{l}|,\mathrm{Re}(Z_{m}^{l}),\mathrm{Im}(Z_{m}^{l})]\right), \\
&a_{m}^{l} = \sigma\!\left(\mathrm{MLP}_{a}^{l}(s_{m}^{l})\right), \quad
\Delta\theta_{m}^{l} = \pi\tanh\!\left(\mathrm{MLP}_{\theta}^{l}(s_{m}^{l})\right), \\
&\hat{Z}_{m}^{l} = \left(a_{m}^{l}\odot G_{m}^{l}\odot Z_{m}^{l}\right)\odot \exp(i\Delta\theta_{m}^{l}), \quad
\hat{z}_{m}^{l}=\mathcal{F}^{-1}(\hat{Z}_{m}^{l}).
\end{aligned}
$}
\end{equation}
Here, $\mathcal{F}(\cdot)$ and $\mathcal{F}^{-1}(\cdot)$ denote the Fourier transform and inverse Fourier transform, respectively.
$\{M_k\}_{k=1}^{K}$ are predefined radial frequency masks, and $c_{m,k}^{l}$ is a learnable complex gain for the $k$-th frequency band.
The radial-band gain captures frequency-selective responses, while the complex channel attention adaptively recalibrates spectral intensity and angular response.

\begin{figure}
    \centering
    \includegraphics[width=\columnwidth]{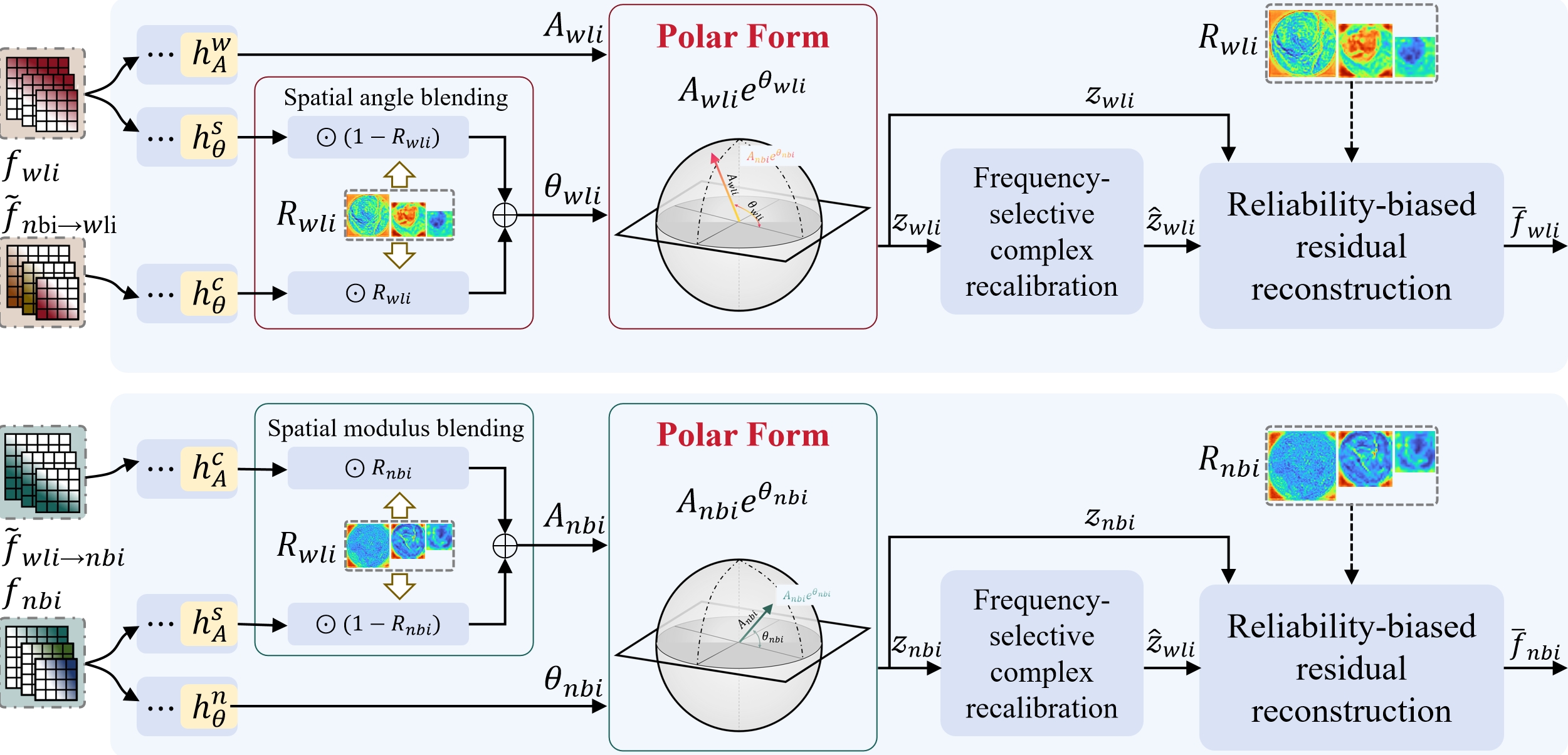}
    \caption{Illustration of the proposed reliability-guided complex spectral fusion (R-CSF) module.}
    \label{fig:CSF}
\vspace{-5mm}
\end{figure}

To avoid over-aggressive spectral rewriting under residual mismatch, we further use a reliability-biased residual reconstruction:
\begin{equation}
\resizebox{0.85\linewidth}{!}{$
\begin{aligned}
\alpha_{m}^{l} &= \sigma\!\left(
h_{g}^{l}\!\left([\mathrm{Re}(z_{m}^{l}),\mathrm{Im}(z_{m}^{l}),\mathrm{Re}(\hat{z}_{m}^{l}),\mathrm{Im}(\hat{z}_{m}^{l})]\right)
+\lambda_{R}R_{m}^{l}
\right), \\
\bar{z}_{m}^{l} &= \alpha_{m}^{l}\odot \hat{z}_{m}^{l} + (1-\alpha_{m}^{l})\odot z_{m}^{l}, \\
\bar{f}_{m}^{l} &= h_{p}^{l}\!\left([\mathrm{Re}(\bar{z}_{m}^{l}),\mathrm{Im}(\bar{z}_{m}^{l})]\right).
\end{aligned}
$}
\end{equation}
Here, $R_{m}^{l}$ denotes $R_{wli}^{l}$ or $R_{nbi}^{l}$ for the corresponding branch, and $\lambda_R$ controls the influence of the reliability prior.
High-reliability regions are allowed to receive stronger spectral recalibration, whereas low-reliability regions preserve more of the original modality-specific complex feature.
The final real-valued features $\{\bar{f}_{\mathrm{wli}}^{l}\}_{l=1}^{L}$ and $\{\bar{f}_{\mathrm{nbi}}^{l}\}_{l=1}^{L}$ are fed into the modality-specific decoders.
Through this reliability-guided polar construction and subsequent Fourier-domain recalibration, R-CSF couples topology-regularized correspondence with role-aware spectral fusion and reduces negative transfer from locally mismatched WLI/NBI observations.

\subsection{Dual-Branch Decoding and Joint Optimization}

The proposed R-CSF module produces enhanced multi-scale features $\{\bar{f}_{wli}^{l}\}_{l=1}^{L}$ and $\{\bar{f}_{nbi}^{l}\}_{l=1}^{L}$ for the two modalities.
We employ two independent FPN-SegFormer-style decoders to aggregate these features through top-down fusion and generate modality-specific lesion probability masks $\hat{y}_{wli}$ and $\hat{y}_{nbi}$.
Auxiliary prediction heads are attached to intermediate scales for deep supervision.

We use branch-wise segmentation supervision for both modalities.
Let $\mathcal{L}_{bce}(\cdot,\cdot)$ and $\mathcal{L}_{dice}(\cdot,\cdot)$ denote the binary cross-entropy loss and Dice loss, respectively.
The segmentation objective is defined as:
\begin{equation}
\resizebox{0.85\linewidth}{!}{$
\begin{aligned}
\mathcal{L}_{seg}
={}&
\sum_{m\in\{\mathrm{wli},\mathrm{nbi}\}}
\Big(
\mathcal{L}_{bce}(\hat{y}_{m},y_{m})
+
\mathcal{L}_{dice}(\hat{y}_{m},y_{m})
\Big) \\
&+
\sum_{l\in\mathcal{A}}\omega_{l}
\sum_{m\in\{\mathrm{wli},\mathrm{nbi}\}}
\Big(
\mathcal{L}_{bce}(\hat{y}_{m}^{\,l},y_{m}^{\,l})
+
\mathcal{L}_{dice}(\hat{y}_{m}^{\,l},y_{m}^{\,l})
\Big).
\end{aligned}
$}
\end{equation}
Here, $\mathcal{A}$ denotes the set of auxiliary prediction scales, $\hat{y}_{m}^{\,l}$ is the auxiliary prediction at scale $l$, $y_{m}^{\,l}$ is the resized ground-truth mask, and $\omega_l$ is the corresponding deep-supervision weight.

Since no dense deformation annotation or explicit pixel-level correspondence is available, R-DRG is optimized through the downstream segmentation objective and alignment-related regularization.
We first impose a smoothness regularization on the stationary velocity field:
\begin{equation}
\resizebox{0.3\linewidth}{!}{$
\begin{aligned}
\mathcal{L}_{smooth}=\|\nabla v\|_{2}^{2}.
\end{aligned}
$}
\end{equation}
To encourage prediction-level semantic consistency between the two modalities, we compare their probability masks after geometric transformation:
\begin{equation}
\resizebox{0.90\linewidth}{!}{$
\begin{aligned}
\mathcal{L}_{prob}
=
\left\|
\mathcal{W}(\hat{y}_{nbi},\phi_{nbi\rightarrow wli})-\hat{y}_{wli}
\right\|_{1}
+
\left\|
\mathcal{W}(\hat{y}_{wli},\phi_{wli\rightarrow nbi})-\hat{y}_{nbi}
\right\|_{1}.
\end{aligned}
$}
\end{equation}
We further introduce a weak lesion-support consistency constraint using the annotated masks:
\begin{equation}
\resizebox{0.90\linewidth}{!}{$
\begin{aligned}
\mathcal{L}_{mask}
=
\left\|
\mathcal{W}(y_{nbi},\phi_{nbi\rightarrow wli})-y_{wli}
\right\|_{1}
+
\left\|
\mathcal{W}(y_{wli},\phi_{wli\rightarrow nbi})-y_{nbi}
\right\|_{1}.
\end{aligned}
$}
\end{equation}
The probability-level and lesion-support constraints regularize task-relevant cross-modal correspondence, but they do not provide dense deformation supervision.
The correspondence reliability maps generated by R-DRG are used only as fusion priors in R-CSF and are not treated as additional supervised targets.

The overall training objective is formulated as:
\begin{equation}
\resizebox{0.85\linewidth}{!}{$
\begin{aligned}
\mathcal{L}
=
\mathcal{L}_{seg}
+
\lambda_{smooth}\mathcal{L}_{smooth}
+
\lambda_{prob}\mathcal{L}_{prob}
+
\lambda_{mask}\mathcal{L}_{mask}.
\end{aligned}
$}
\end{equation}
Here, $\lambda_{smooth}$, $\lambda_{prob}$, and $\lambda_{mask}$ balance the different regularization terms.
In this way, R-DRG and R-CSF are jointly optimized under the final segmentation objective, allowing the learned deformation to provide task-relevant correspondence while the reliability-guided spectral fusion selectively exploits trustworthy WLI/NBI complementary cues.

\section{Experiment}\label{sec_experiment}

\subsection{Datasets and Paired Sample Construction}
\label{subsec_dataset}

We evaluated our method on two WLI/NBI endoscopic datasets: the private FAHSYSU dataset \cite{jiepy_eClinicalMedicine,jiepy_AoM} and the public PICCOLO dataset \cite{PICCOLO}. Since WLI and NBI images were acquired separately without pixel-wise correspondence, we constructed lesion-matched but unregistered pairs, each comprising a WLI image, an NBI image, and their respective binary lesion masks. Candidate pairs were independently reviewed by three clinicians (two rhinologists with over 10 years of experience each and one resident physician with 5 years of experience). Pair validity required the same patient and lesion identity, verified using available examination identifiers and lesion-location records. Disagreements were resolved through consensus review. For FAHSYSU, pathological records, including biopsy results, were additionally reviewed to confirm lesion identity and diagnosis.

As each lesion could be imaged multiple times under different viewpoints or imaging conditions, all clinically valid WLI--NBI combinations were retained without selection based on alignment quality. The resulting pair lists were fixed throughout training and evaluation, with no re-pairing or alignment-based selection. Each pair was treated as a distinct dual-modality input and contributed equally to pair-level evaluation. Even when different pairs shared a source image, their distinct counterpart observations could produce different cross-modal correspondences and fusion responses.

The resulting FAHSYSU (Paired) dataset contained 14,529 pairs (11,512 training and 3,017 testing), while PICCOLO (Paired) comprised 7,362 pairs (5,529 training and 1,833 testing). Patient-level splitting ensured that all images and pairs from the same patient, including those associated with the same lesion, were assigned exclusively to either training or testing, preventing patient- and lesion-level data leakage. All methods were evaluated on the same test splits, with paired-input methods using identical fixed WLI--NBI pairs.

Importantly, the unit of evaluation in this study is a lesion-matched WLI–NBI pair rather than an independent patient or source image. Since multiple observations of the same lesion may be available, distinct clinically valid pairs can share an original image. Such reuse is confined within the patient-disjoint data partitions and does not introduce overlap between training and testing. Retaining these pairs allows evaluation under diverse acquisition conditions and cross-modal discrepancies. All competing methods are evaluated using the same fixed pair lists and pair-level protocol, ensuring consistent comparisons within the constructed benchmark. Nevertheless, the resulting performance estimates are pair-weighted and should not be interpreted as equally weighted patient-level estimates.

\subsection{Implementation Details}

All experiments were implemented in PyTorch on four NVIDIA RTX 3090 GPUs. 
All RGB images were resized to $512\times512$ and normalized with ImageNet statistics, without additional online augmentation. 
We used a shared ConvNeXtV2 encoder with lightweight modality-specific adapters. 
The proposed framework uses $L=3$ pyramid feature levels. 
For the $512\times512$ input resolution, the extracted feature maps have spatial resolutions of $64\times64$, $32\times32$, and $16\times16$, respectively. 
The stationary velocity field is predicted only at $16\times 16$ resolution and subsequently resized to other pyramid levels for multi-scale deformation warping. 
No independent correspondence reliability estimation module is employed in the current implementation.
The network was optimized using AdamW with a weight decay of $1\times10^{-4}$. 
Automatic mixed precision and distributed data parallel training were adopted for efficiency. 
Different learning rates were used for the backbone, adapters, and other modules, with epoch-wise multiplicative decay. 
After a predefined epoch, the shared backbone and feature projection layers were frozen, while the adapters and downstream modules were further optimized.

\subsection{Evaluation Metrics}

We evaluate segmentation performance using five standard region-level metrics, including mean Intersection over Union (mIoU), Dice Similarity Coefficient (DSC), Pixel Accuracy (PA), Recall, and Precision~\cite{jiepy_TCSVT,xiang2024lightweight}. Pixel-wise class labels are obtained by applying argmax to the softmax outputs.
For each test sample, the WLI prediction is evaluated against the WLI ground-truth mask and the NBI prediction is evaluated against the NBI ground-truth mask.
For each modality-specific mask, let $TP$, $FP$, $FN$, and $TN$ denote the numbers of true-positive, false-positive, false-negative, and true-negative pixels, respectively, with lesion pixels treated as the positive class. For binary segmentation, mIoU is averaged over foreground and background classes ($k=2$), and the metrics are defined as
\begin{equation*}
\resizebox{0.65\linewidth}{!}{$
\begin{aligned}
\mathrm{mIoU} &=
\frac{1}{k}\sum_{i=1}^{k}
\frac{TP_i}{TP_i+FP_i+FN_i}, \\
\mathrm{DSC} &=
\frac{2TP}{2TP+FP+FN}, \\
\mathrm{PA} &=
\frac{TP+TN}{TP+TN+FP+FN}, \\
\mathrm{Recall} &=
\frac{TP}{TP+FN}, 
\mathrm{Precision} =
\frac{TP}{TP+FP}.
\end{aligned}
$}
\end{equation*}
Here, mIoU and DSC measure region overlap, Recall and Precision reflect the balance between false negatives and false positives, and PA is reported as a complementary metric. 
For each method, quantitative results are reported as the mean and standard deviation of these per-run dataset-level averages over four independent training runs with different random seeds.

To further assess boundary quality, we additionally report the 95\% Hausdorff Distance ($\mathrm{HD}_{95}$), Average Symmetric Surface Distance (ASSD)~\cite{ADFNet,HD95_ASSD}, and Boundary IoU (BIoU)~\cite{BIOU}. These are used to evaluate contour discrepancy and boundary-region overlap, which are particularly relevant when residual cross-modal mismatch exists in paired but unregistered WLI--NBI images. HD95 and ASSD are computed in pixels on the resized $512\times512$ masks, and BIoU is calculated using a boundary band radius of $r=2$ pixels.

\subsection{Compared Methods}

We compare our method with ten representative baselines, including UMF-SegNet \cite{UMFSegNet}, DFormer \cite{DFormer}, FTransUNet \cite{FTransUNet}, DCRS \cite{DCRS}, VI-ReID \cite{VI-ReID}, ShapeConv \cite{ShapeConv}, MCSNet \cite{MCSNet}, MFNet \cite{MFNet}, UMSCS \cite{UMSCS}, and ADFNet \cite{ADFNet}. The compared methods span three relevant categories, namely misalignment-aware multimodal learning, generic dual-stream fusion, and registration- or correspondence-related approaches. Among them, ADFNet and UMSCS are the most closely related baselines, as they explicitly address cross-modal discrepancy through disentanglement or cross-modal learning. However, most compared methods still rely on spatial-domain or real-valued feature interaction, and do not explicitly model the frequency-level complementarity between WLI and NBI. For fair comparison, all methods were trained and evaluated using the same data splits, input resolution, preprocessing pipeline, and optimization settings whenever applicable. For baselines originally developed for other multimodal scenarios such as RGB-D and remote sensing, we preserved their original multimodal learning paradigms while adapting their inputs to the WLI and NBI modalities. For methods originally designed to produce only a single segmentation output, we adopted a shared-weight, modality-swapped training strategy, where paired WLI/NBI images were alternately processed in the orders (WLI, NBI) and (NBI, WLI), with each prediction supervised by the lesion mask of the first (target) modality. During inference, the same model performed two forward passes with swapped input orders to generate the WLI and NBI segmentation masks, respectively, while preserving cross-modal feature interaction in each pass. Therefore, all compared methods were evaluated under the same dual-modality segmentation objective, regardless of whether their original architectures explicitly supported dual-output prediction.

\begin{table*}[!t]
\caption{Comparison of methods on PICCOLO and FAHSYSU (mean $\pm$ standard deviation, n=4).}
\label{Table:PICCOLO_FAHSYSU}
\belowrulesep=0pt
\aboverulesep=0pt
\centering
\renewcommand{\arraystretch}{1.2}
\setlength{\tabcolsep}{2.2pt}
\begin{tabular*}{\textwidth}{@{\extracolsep{\fill}}c|l|ccccc}
\toprule
Dataset & Methods & mIoU (\%) $\uparrow$ & DSC (\%) $\uparrow$ & PA (\%) $\uparrow$ & Recall (\%) $\uparrow$ & Precision (\%) $\uparrow$ \\
\hline
\multirow{11}{*}{\begin{tabular}[c]{@{}c@{}}PICCOLO\\(Paired)\end{tabular}}
& UMF-SegNet \cite{UMFSegNet} (KBS 2025) & 62.61\std{0.44} & 63.26\std{0.50} & 83.09\std{0.69} & 76.43\std{0.72} & 53.96\std{0.64} \\
& DFormer \cite{DFormer} (ICLR 2024) & \second{80.62\std{0.51}} & 82.43\std{0.55} & 93.57\std{0.71} & \second{89.69\std{0.75}} & 76.26\std{0.76} \\
& FTransUNet \cite{FTransUNet} (TGRS 2024) & 63.59\std{0.25} & 65.96\std{0.50} & 83.72\std{0.61} & 78.40\std{0.55} & 56.93\std{0.68} \\
& DCRS \cite{DCRS} (Bioinformatics 2024) & 77.31\std{0.93} & 77.79\std{0.42} & {93.20\std{1.04}} & 81.96\std{0.73} & 74.03\std{0.48} \\
& VI-ReID \cite{VI-ReID} (IF 2024) & 72.48\std{0.56} & 75.22\std{0.59} & 91.66\std{1.10} & 72.41\std{0.89} & 78.26\std{0.73} \\
& ShapeConv \cite{ShapeConv} (ICCV 2021) & 57.91\std{0.24} & 59.09\std{0.35} & 81.27\std{0.40} & 70.24\std{0.28} & 51.00\std{0.50} \\
& MCSNet \cite{MCSNet} (TNNLS 2024) & 76.77\std{0.89} & 78.34\std{0.31} & 91.38\std{0.91} & 84.70\std{0.43} & 72.87\std{0.44} \\
& MFNet \cite{MFNet} (TGRS 2025) & 62.75\std{0.76} & 62.61\std{0.56} & 83.13\std{0.63} & 63.86\std{0.83} & 61.41\std{0.75} \\
& ADFNet \cite{ADFNet} (MIA 2026) & 82.05\std{0.81} & \second{82.48\std{0.30}} & \best{94.67\std{0.43}} & 84.90\std{0.40} & \second{80.20\std{0.44}} \\
& UMSCS \cite{UMSCS} (IJCV 2025) & 78.87\std{0.60} & 81.91\std{0.28} & 92.94\std{0.45} & 85.01\std{0.40} & 79.02\std{0.38} \\
& Ours & \best{83.54\std{0.31}} & \best{84.78\std{0.36}} & \second{94.49\std{0.26}} & \best{89.85\std{0.75}} & \best{80.25\std{0.24}} \\
\hline
\multirow{11}{*}{\begin{tabular}[c]{@{}c@{}}FAHSYSU\\(Paired)\end{tabular}}
& UMF-SegNet \cite{UMFSegNet} (KBS 2025) & 73.18\std{0.92} & 74.48\std{0.42} & 89.42\std{1.05} & 78.43\std{0.78} & 70.90\std{0.41} \\
& DFormer \cite{DFormer} (ICLR 2024) & 72.64\std{0.51} & 73.59\std{0.31} & 89.87\std{0.62} & 73.50\std{0.39} & 73.68\std{0.47} \\
& FTransUNet \cite{FTransUNet} (TGRS 2024) & 75.20\std{0.76} & 77.09\std{0.51} & 90.11\std{0.58} & 80.51\std{0.86} & 73.94\std{0.60} \\
& DCRS \cite{DCRS} (Bioinformatics 2024) & 77.08\std{0.38} & 78.26\std{0.29} & 91.89\std{0.54} & 78.82\std{0.42} & 77.71\std{0.39} \\
& VI-ReID \cite{VI-ReID} (IF 2024) & 75.35\std{0.34} & 77.08\std{0.28} & 91.41\std{0.45} & 75.14\std{0.43} & 79.13\std{0.36} \\
& ShapeConv \cite{ShapeConv} (ICCV 2021) & 74.63\std{0.36} & 75.37\std{0.30} & 90.87\std{0.58} & 75.21\std{0.46} & 75.54\std{0.40} \\
& MCSNet \cite{MCSNet} (TNNLS 2024) & 75.16\std{0.29} & 77.03\std{0.25} & 90.79\std{0.41} & 80.57\std{0.32} & 73.78\std{0.38} \\
& MFNet \cite{MFNet} (TGRS 2025) & 72.96\std{0.33} & 75.95\std{0.25} & 88.41\std{0.47} & \second{82.18\std{0.35}} & 70.60\std{0.34} \\
& ADFNet \cite{ADFNet} (MIA 2026) & \second{79.72\std{0.27}} & \second{80.72\std{0.19}} & \second{92.96\std{0.33}} & 80.79\std{0.28} & \second{80.66\std{0.26}} \\
& UMSCS \cite{UMSCS} (IJCV 2025) & 79.09\std{0.44} & 80.30\std{0.27} & 92.88\std{0.32} & 81.01\std{0.40} & 79.60\std{0.36} \\
& Ours & \best{81.08\std{0.23}} & \best{82.37\std{0.17}} & \best{93.40\std{0.30}} & \best{83.68\std{0.25}} & \best{81.11\std{0.22}} \\
\bottomrule
\end{tabular*}
\vspace{-5mm}

\end{table*}

\subsubsection{Quantitative Analysis}

Table~\ref{Table:PICCOLO_FAHSYSU} reports the quantitative comparison on PICCOLO and FAHSYSU. Overall, our method achieves the best region-level segmentation performance on both datasets. On PICCOLO, it obtains the highest mIoU, DSC, Recall, and Precision, reaching 83.54\%, 84.78\%, 89.85\%, and 80.25\%, respectively, while achieving the second-best PA of 94.49\%. Compared with the strongest competing results, our method improves mIoU by 2.24\% over DFormer~\cite{DFormer} and DSC by 2.30\% over ADFNet~\cite{ADFNet}. On FAHSYSU, our method achieves the best results across all five metrics, with 81.08\% mIoU, 82.37\% DSC, 93.40\% PA, 83.68\% Recall, and 81.11\% Precision. Compared with ADFNet, it improves mIoU, DSC, PA, and Precision by 1.36\%, 1.65\%, 0.44\%, and 0.45\%, respectively, and improves Recall by 1.50\% over the second-best MFNet~\cite{MFNet}.
The simultaneous improvement in Recall and Precision indicates that the proposed method does not simply enlarge foreground predictions to improve lesion coverage, but better balances false negatives and false positives. This is particularly important for endoscopic lesion segmentation, where low-contrast regions and ambiguous boundaries often lead to missed lesions or over-segmentation. Compared with alignment-aware methods such as ADFNet~\cite{ADFNet} and UMSCS~\cite{UMSCS}, generic multimodal fusion methods~\cite{UMFSegNet,DFormer,FTransUNet,ShapeConv,MCSNet,MFNet}, and registration-based DCRS~\cite{DCRS}, our method more explicitly addresses both cross-modal geometric mismatch and modality complementarity.

\begin{figure}
    \centering
    \includegraphics[width=\columnwidth]{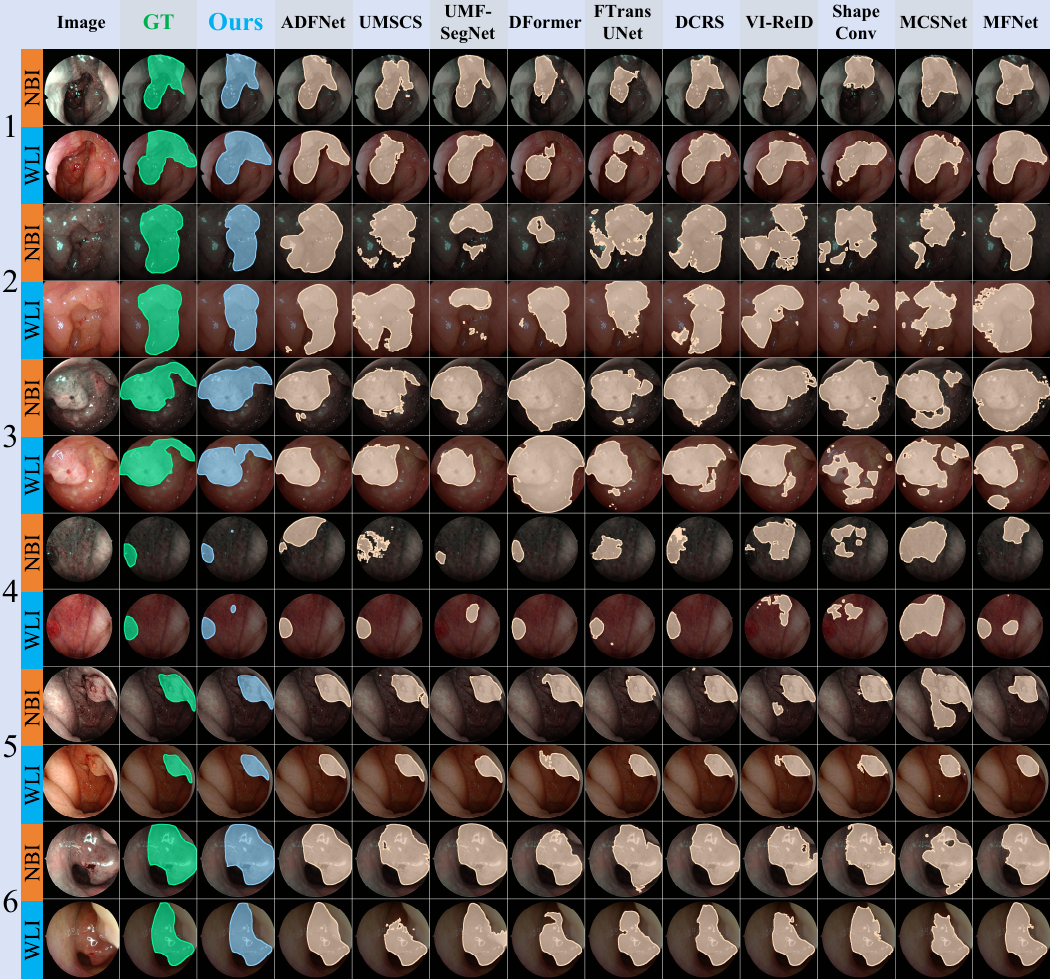}
    \caption{Qualitative comparison on FAHSYSU (Paired).}
    \label{fig:compare_SYSU}
\vspace{-5mm}
\end{figure} 

These results support the effectiveness of coupling diffeomorphic feature registration with complex spectral fusion. The registration module provides topology-preserving cross-modal correspondence before fusion, while reliability-guided polar construction and subsequent Fourier-domain recalibration allow WLI-derived appearance cues and NBI-derived structural information to be integrated in a controlled manner. Therefore, the proposed framework is better suited to paired but unregistered WLI/NBI lesion segmentation than conventional real-valued spatial-domain fusion or alignment alone.

\begin{figure}
    \centering
    \includegraphics[width=\columnwidth]{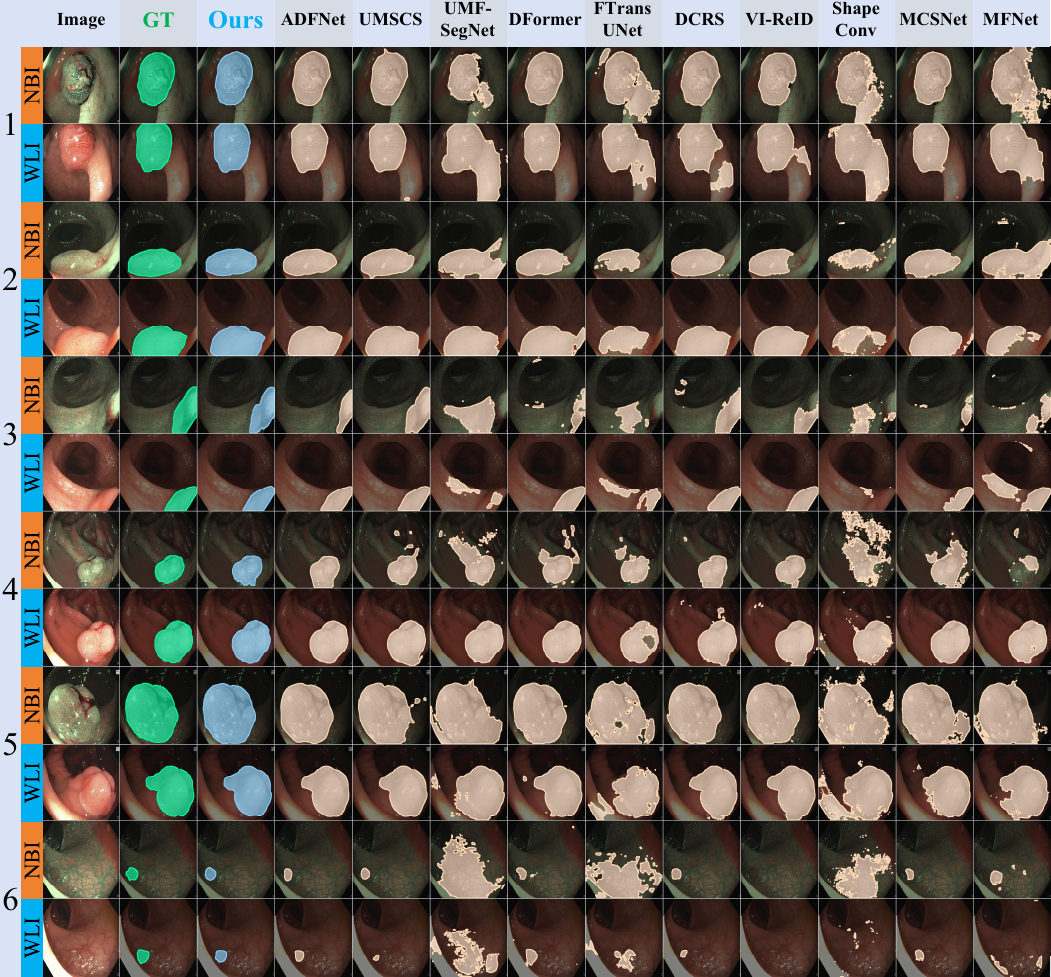}
    \caption{Qualitative comparison on PICCOLO (Paired).}
    \label{fig:compare_PICCOLO}
\vspace{-5mm}
\end{figure} 

\subsubsection{Qualitative Analysis}

As shown in Fig.~\ref{fig:compare_SYSU} and Fig.~\ref{fig:compare_PICCOLO}, our method produces more complete lesion masks and cleaner boundaries than the competing methods, especially in examples with low contrast, ambiguous contours, irregular lesion morphology, or noticeable WLI/NBI viewpoint differences. Several baselines tend to miss subtle lesion regions or incorrectly include surrounding normal mucosa, resulting in fragmented masks or over-expanded predictions. In contrast, our method better preserves lesion continuity while suppressing false-positive responses. This visual trend is consistent with the quantitative results in Table~\ref{Table:PICCOLO_FAHSYSU}, where our method achieves the best mIoU, DSC, Recall, and Precision on both datasets, indicating improved lesion coverage without sacrificing prediction reliability.

The qualitative advantage is particularly evident when one modality provides insufficient or ambiguous information, such as examples 1, 2, 4, and 6 in Fig.~\ref{fig:compare_SYSU} and examples 3, 4, and 6 in Fig.~\ref{fig:compare_PICCOLO}. By reducing cross-modal mismatch before fusion and then performing frequency-aware interaction between WLI appearance cues and NBI structural details, the proposed framework can recover weakly contrasted lesion regions, maintain coherent boundaries, and avoid unreliable enhancement in mismatched normal tissues. These results further demonstrate the effectiveness of registration-grounded complex spectral fusion for paired but unregistered WLI/NBI lesion segmentation.

\subsection{Ablation Study}

To comprehensively analyze the effectiveness of the proposed framework, we conduct two groups of ablation studies. First, we evaluate the contribution of the main modules, including the diffeomorphic feature registration module and the complex spectral fusion module. Second, we compare the proposed dual-modality setting with single-modality variants to verify the necessity of exploiting WLI/NBI complementarity.

\subsubsection{Module Ablation}

To validate the contribution of each key component, we construct four settings: a baseline without R-DRG and R-CSF, a variant using R-CSF without registration grounding, a variant using R-DRG with real-domain fusion, and the complete model with both R-DRG and R-CSF.
For the R-CSF-only variant, the reliability guidance is disabled so that it evaluates the effect of complex spectral fusion without registration grounding.
All variants share the same encoder, decoder, and training protocol, so that the performance differences mainly reflect the effects of reliability-aware correspondence grounding and complex spectral fusion.

As shown in Table~\ref{Table:ablation_experiment}, R-DRG and R-CSF provide complementary benefits.
On PICCOLO, the baseline without both modules achieves 78.92\% mIoU and 81.90\% DSC.
Using R-CSF alone improves mIoU to 79.87\%, but does not improve DSC, indicating that complex spectral fusion may be less stable without reliable cross-modal correspondence.
Using R-DRG with real-domain fusion improves DSC to 82.76\%, suggesting that topology-regularized correspondence is beneficial even without complex-domain interaction.
Combining R-DRG and R-CSF yields the best overall performance, reaching 83.54\% mIoU and 84.78\% DSC.

The same trend is observed on FAHSYSU, where the complete model improves mIoU and DSC from 78.02\%/80.53\% to 81.08\%/82.37\%.
Although the R-CSF-only variant gives a slightly higher Recall on FAHSYSU, the complete model achieves better mIoU, DSC, PA, and Precision, indicating a better balance between lesion coverage and false-positive control.
These results show that R-DRG provides a reliable geometric and correspondence basis, while R-CSF further exploits WLI/NBI complementarity through reliability-guided complex-domain fusion.

\begin{table}[!h]
    \belowrulesep=0pt
    \aboverulesep=0pt
    \centering
    \caption{Module ablation results on PICCOLO and FAHSYSU.}
    \label{Table:ablation_experiment}
    \renewcommand{\arraystretch}{1.25}
    \setlength{\tabcolsep}{2.2pt}
    \resizebox{\columnwidth}{!}{
    \begin{tabular}{c|cc|ccccc}
    \toprule
    \multirow{2}{*}{Dataset} &
    \multicolumn{2}{c|}{Ablation Settings} &
    \multirow{2}{*}{$\text{mIoU}$ $(\%)$} &
    \multirow{2}{*}{$\text{DSC}$ $(\%)$} &
    \multirow{2}{*}{$\text{PA}$ $(\%)$} &
    \multirow{2}{*}{$\text{Recall}$ $(\%)$} &
    \multirow{2}{*}{$\text{Precision}$ $(\%)$} \\
    \cmidrule(lr){2-3}
    & R-DRG & R-CSF \\
    \hline
    \multirow{4}{*}{\makecell{PICCOLO\\(Paired)}}
    & \usym{1F5F4} & \usym{1F5F4} & 78.92\std{0.62} & 81.90\std{0.31} & 92.00\std{0.55} & 85.03\std{0.54} & 79.00\std{0.33} \\
    & \usym{1F5F4} & \checkmark & 79.87\std{0.21} & 81.76\std{0.44} & 92.20\std{0.34} & 85.11\std{0.53} & 78.67\std{0.67} \\
    & \checkmark & \usym{1F5F4} & 79.25\std{0.41} & 82.76\std{0.34} & 92.53\std{0.34} & 86.10\std{0.53} & 79.67\std{0.44} \\
    & \checkmark & \checkmark & \best{83.54\std{0.31}} & \best{84.78\std{0.36}} & \best{94.49\std{0.26}} & \best{89.85\std{0.75}} & \best{80.25\std{0.24}} \\
    \hline
    \multirow{4}{*}{\makecell{FAHSYSU\\(Paired)}}
    & \usym{1F5F4} & \usym{1F5F4} & 78.02\std{0.22} & 80.53\std{0.31} & 90.87\std{0.45} & 81.45\std{0.54} & 79.63\std{0.33} \\
    & \usym{1F5F4} & \checkmark & 79.97\std{0.21} & 81.59\std{0.16} & 93.02\std{0.25} & \best{83.79\std{0.22}} & 79.51\std{0.24} \\
    & \checkmark & \usym{1F5F4} & 79.63\std{0.22} & 81.36\std{0.16} & 92.93\std{0.24} & 82.35\std{0.23} & 80.39\std{0.22} \\
    & \checkmark & \checkmark & \best{81.08\std{0.23}} & \best{82.37\std{0.17}} & \best{93.40\std{0.30}} & 83.68\std{0.25} & \best{81.11\std{0.22}} \\
    \bottomrule
    \end{tabular}
    }
\vspace{-5mm}
    
\end{table}
 
\subsubsection{Registration Baseline Comparison}

To evaluate the proposed registration strategy, we replace R-DRG with TransMorph \cite{TransMorph} while retaining the same feature warping, real-valued fusion, and segmentation decoders. As shown in Table~\ref{tab:registration_fusion_baseline}, R-DRG achieves modest improvements over TransMorph, although their mIoU on PICCOLO is nearly identical. In contrast, incorporating R-CSF yields more substantial gains across both datasets, suggesting that spatial alignment alone is insufficient to fully exploit cross-modal complementarity. These findings highlight the benefits of combining registration with reliability-guided, role-aware complex spectral fusion.
\begin{table}[t]
\centering
\caption{Comparison with a registration--fusion--segmentation baseline on FAHSYSU and PICCOLO.}
\label{tab:registration_fusion_baseline}

\footnotesize
\setlength{\tabcolsep}{3pt}
\renewcommand{\arraystretch}{1.15}

\resizebox{\columnwidth}{!}{%
\begin{tabular}{llcccc}
\toprule
\multirow{2}{*}{Registration} &
\multirow{2}{*}{Fusion} &
\multicolumn{2}{c}{FAHSYSU (Paired)} &
\multicolumn{2}{c}{PICCOLO (Paired)} \\
\cmidrule(lr){3-4}
\cmidrule(lr){5-6}
& & mIoU (\%) $\uparrow$ & DSC (\%) $\uparrow$
  & mIoU (\%) $\uparrow$ & DSC (\%) $\uparrow$ \\
\midrule

TransMorph
& Real-valued
& 78.77\std{0.30} 
& 81.02\std{0.32}
& 79.22\std{0.40} 
& 82.17\std{0.37} \\

R-DRG (Ours)
& Real-valued
& 79.63\std{0.22}
& 81.36\std{0.16}
& 79.25\std{0.41}
& 82.76\std{0.34} \\

R-DRG (Ours)
& R-CSF (Ours)
& 81.08\std{0.23}
& 82.37\std{0.17}
& 83.54\std{0.31}
& 84.78\std{0.36} \\

\bottomrule
\end{tabular}%
}
\vspace{-4mm}
\end{table}

\subsubsection{Modality Ablation}

We further compare the proposed dual-modality setting with WLI-only and NBI-only variants to verify the contribution of multimodal information. As shown in Table~\ref{tab:single_modal}, using both modalities consistently improves all metrics on both datasets. On FAHSYSU, the dual-modality model improves mIoU/DSC by 3.28\%/2.59\% over WLI-only and 2.87\%/2.23\% over NBI-only. On PICCOLO, it improves mIoU/DSC by 6.06\%/3.13\% and 5.81\%/2.87\%, respectively. These gains indicate that WLI and NBI provide complementary cues, improving lesion coverage while maintaining cleaner predictions.

\begin{table}[!h]
\centering
\caption{Modality ablation results with single-modality and dual-modality settings.}
\label{tab:single_modal}
\scriptsize
\setlength{\tabcolsep}{1.8pt}
\renewcommand{\arraystretch}{0.88}
\resizebox{\columnwidth}{!}{
\begin{tabular}{c c c c c c c c}
\toprule
Dataset & WLI & NBI & mIoU & DSC & PA & Recall & Precision \\
\midrule
\multirow{3}{*}{\makecell{FAHSYSU\\(Paired)}}
& \checkmark &  & 77.80\sstd{0.33} & 79.78\sstd{0.22} & 89.94\sstd{0.38} & 82.50\sstd{0.26} & 77.24\sstd{0.34} \\
&  & \checkmark & 78.21\sstd{0.51} & 80.14\sstd{0.27} & 90.12\sstd{0.56} & 82.23\sstd{0.44} & 78.15\sstd{0.33} \\
& \checkmark & \checkmark & \best{81.08\sstd{0.23}} & \best{82.37\sstd{0.17}} & \best{93.40\sstd{0.30}} & \best{83.68\sstd{0.25}} & \best{81.11\sstd{0.22}} \\
\midrule
\multirow{3}{*}{\makecell{PICCOLO\\(Paired)}}
& \checkmark &  & 76.80\sstd{0.33} & 81.65\sstd{0.18} & 90.28\sstd{0.28} & 84.74\sstd{0.26} & 78.77\sstd{0.24} \\
&  & \checkmark & 77.05\sstd{0.31} & 81.91\sstd{0.17} & 90.57\sstd{0.26} & 85.06\sstd{0.24} & 78.99\sstd{0.23} \\
& \checkmark & \checkmark & \best{83.54\sstd{0.31}} & \best{84.78\sstd{0.36}} & \best{94.49\sstd{0.26}} & \best{89.85\sstd{0.75}} & \best{80.25\sstd{0.24}} \\
\bottomrule
\end{tabular}
}
\vspace{-5mm}
\end{table}

\subsection{Effect of Role Assignment in Polar Representations}

To investigate the contribution of modality-role assignment within the proposed complex fusion framework, we compare the proposed configuration with a real-valued fusion baseline and a reversed polar-role configuration.
Specifically, in the Real-valued Fusion setting, the complex magnitude--phase construction and subsequent complex spectral modulation are removed; at each pyramid level, the in-branch feature and its R-DRG-aligned cross-modal counterpart are concatenated along the channel dimension and mapped through a learnable real-valued projection. In the Reversed Role setting, the same complex-fusion pipeline is retained, but the modality roles are swapped relative to the proposed assignment, such that NBI-derived cues serve as the magnitude source and WLI-derived cues serve as the phase source. The Proposed Role follows the assignment defined in Eq.~\ref{eq:role_assignment}, where WLI-derived cues provide the magnitude-related appearance information and NBI-derived cues provide the phase-related structural information. Except for the fusion operator or magnitude--phase role assignment under evaluation, the encoder, R-DRG module, decoders, loss functions, and training protocol are kept unchanged across the three settings.
As shown in Table~\ref{tab:role_assignment}, the configuration that uses WLI-derived features to parameterize the magnitude component and NBI-derived features to parameterize the phase component consistently achieves the best performance on both datasets. Compared with real-valued fusion, it improves mIoU/DSC by 1.45\%/1.01\% on FAHSYSU and 3.61\%/2.02\% on PICCOLO, demonstrating the advantage of complex spectral modeling over direct real-domain interaction.

Although the reversed configuration also outperforms real-valued fusion in most metrics, it remains inferior to the proposed one, especially in mIoU, DSC, and Recall. This comparison empirically supports assigning WLI-derived cues to the learned spatial modulus and NBI-derived cues to the learned spatial angle. Rather than implying a direct transfer of image-level Fourier magnitude and phase, these findings demonstrate the effectiveness of the modality-asymmetric polar parameterization within the proposed fusion architecture. Together with the subsequent Fourier-domain recalibration, the results support the benefit of role-aware complex spectral fusion over the tested alternative fusion strategies.

\begin{table}[!htbp]
\centering
\caption{Effect of modality-role assignment in learned spatial polar representations. Real-valued: aligned feature concatenation with real-valued projection; Reversed: NBI$\rightarrow$magnitude and WLI$\rightarrow$phase; Proposed: WLI$\rightarrow$magnitude and NBI$\rightarrow$phase.}
\label{tab:role_assignment}
\footnotesize
\setlength{\tabcolsep}{3pt}
\renewcommand{\arraystretch}{0.95}
\resizebox{\columnwidth}{!}{
\begin{tabular}{llccc}
\toprule
Dataset & Metric 
& Real-valued Fusion
& Reversed Role
& Proposed Role \\
\midrule
\multirow{5}{*}{\rotatebox[origin=c]{90}{\fontsize{8pt}{7.5pt}\selectfont\shortstack{FAHSYSU\\{\fontsize{5.8pt}{6.5pt}\selectfont (Paired)}}}}
& mIoU      & 79.63{\scriptsize $\pm$0.22} & 80.05{\scriptsize $\pm$0.56} & 81.08{\scriptsize $\pm$0.23} \\
& DSC       & 81.36{\scriptsize $\pm$0.16} & 81.44{\scriptsize $\pm$0.43} & 82.37{\scriptsize $\pm$0.17} \\
& PA        & 92.93{\scriptsize $\pm$0.24} & 92.88{\scriptsize $\pm$0.21} & 93.40{\scriptsize $\pm$0.30} \\
& Recall    & 82.35{\scriptsize $\pm$0.23} & 82.50{\scriptsize $\pm$0.59} & 83.68{\scriptsize $\pm$0.25} \\
& Precision & 80.39{\scriptsize $\pm$0.22} & 80.40{\scriptsize $\pm$0.63} & 81.11{\scriptsize $\pm$0.22} \\
\midrule
\multirow{5}{*}{\rotatebox[origin=c]{90}{\fontsize{8pt}{7.5pt}\selectfont\shortstack{PICCOLO\\{\fontsize{5.8pt}{6.5pt}\selectfont (Paired)}}}}
& mIoU      & 79.25{\scriptsize $\pm$0.41} & 81.06{\scriptsize $\pm$0.62} & 83.54{\scriptsize $\pm$0.31} \\
& DSC       & 82.76{\scriptsize $\pm$0.34} & 83.26{\scriptsize $\pm$0.31} & 84.78{\scriptsize $\pm$0.36} \\
& PA        & 92.53{\scriptsize $\pm$0.34} & 93.00{\scriptsize $\pm$0.28} & 94.49{\scriptsize $\pm$0.26} \\
& Recall    & 86.10{\scriptsize $\pm$0.53} & 87.17{\scriptsize $\pm$0.37} & 89.85{\scriptsize $\pm$0.75} \\
& Precision & 79.67{\scriptsize $\pm$0.44} & 79.68{\scriptsize $\pm$0.47} & 80.25{\scriptsize $\pm$0.24} \\
\bottomrule
\end{tabular}}
\vspace{-4mm}

\end{table}

\subsection{Boundary-Sensitive Evaluation Analysis}
Beyond region-level metrics, we evaluate boundary quality using $\mathrm{HD}_{95}$, ASSD, and $\mathrm{BIoU}_{r=2}$. As shown in Table~\ref{tab:hd95_assd_biou}, our method achieves the best ASSD and BIoU on PICCOLO, and ranks second in $\mathrm{HD}_{95}$ only behind ADFNet. On FAHSYSU, it obtains the best results across all three boundary metrics. These results indicate that the proposed framework improves not only region overlap but also contour-level lesion delineation. This benefit is consistent with our design. R-DRG mitigates boundary drift caused by cross-modal mismatch, while R-CSF enhances the interaction between WLI appearance cues and NBI structural details.

\begin{table}[!htbp]
    \caption{Boundary-sensitive comparison.}
    \label{tab:hd95_assd_biou}
    \centering
    \belowrulesep=0pt
    \aboverulesep=0pt
    \renewcommand{\arraystretch}{1.08}
    \setlength{\tabcolsep}{3.0pt}
    \footnotesize
    \begin{tabular}{c|l|ccc}
        \toprule
        Dataset & Methods & $\mathrm{HD}_{95}$ $\downarrow$ & ASSD $\downarrow$ & BIoU$_{r=2}$ (\%) $\uparrow$ \\
        \hline
        \multirow{11}{*}{\rotatebox[origin=c]{0}{\makecell{PICCOLO\\(Paired)}}}
        & UMF-SegNet & 168.89 & 56.78 & 10.54 \\
        & DFormer & 175.61 & 41.56 & 11.47 \\
        & FTransUNet & 159.40 & 45.55 & 8.58 \\
        & DCRS & 134.58 & 37.91 & 19.83 \\
        & VI-ReID & 121.51 & 31.27 & 21.51 \\
        & ShapeConv & 170.35 & 47.49 & 6.92 \\
        & MCSNet & 174.91 & 43.22 & 7.30 \\
        & MFNet & 156.27 & 50.26 & 13.30 \\
        & ADFNet & \best{89.05} & \second{22.60} & \second{22.44} \\
        & UMSCS & 101.40 & 23.52 & 19.63 \\
        & Ours & \second{98.78} & \best{20.26} & \best{24.32} \\
        \hline
        \multirow{11}{*}{\rotatebox[origin=c]{0}{\makecell{FAHSYSU\\(Paired)}}}
        & UMF-SegNet & 104.03 & 32.55 & 9.86 \\
        & DFormer & 107.44 & 34.16 & 8.52 \\
        & FTransUNet & 100.47 & 30.75 & 6.99 \\
        & DCRS & 89.45 & 25.04 & 11.71 \\
        & VI-ReID & 96.03 & 27.91 & 10.53 \\
        & ShapeConv & 114.43 & 36.53 & 5.24 \\
        & MCSNet & 113.46 & 34.42 & 5.70 \\
        & MFNet & 104.76 & 31.31 & 10.49 \\
        & ADFNet & \second{75.98} & 22.37 & \second{15.11} \\
        & UMSCS & 76.53 & \second{21.58} & 12.62 \\
        & Ours & \best{69.04} & \best{19.90} & \best{15.96} \\
        \bottomrule
    \end{tabular}
\end{table}

\subsection{Cross-Modal Feature Registration Analysis}

To quantitatively evaluate the effectiveness of the proposed registration module, we assess cross-modal correspondence before and after warping on the entire FAHSYSU and PICCOLO test sets. Two complementary metrics are adopted: cosine similarity, which measures the agreement between corresponding WLI and NBI feature representations, and DSC, which evaluates the spatial overlap between the annotated lesion masks of the target and moving modalities before and after applying the estimated deformation field. Both metrics are computed for NBI$\rightarrow$WLI and WLI$\rightarrow$NBI registration and averaged over all test pairs.

As shown in Table~\ref{tab:registration_quality}, both cosine similarity and DSC consistently improve after warping across both datasets and registration directions, demonstrating enhanced cross-modal feature correspondence and lesion-region alignment. Notably, the substantial DSC improvements on PICCOLO, despite its relatively low initial spatial overlap, indicate that R-DRG effectively mitigates pronounced cross-modal misalignment. These results provide complementary quantitative evidence that the predicted deformation fields establish task-relevant spatial correspondence for subsequent multimodal fusion.

\begin{table}[t]
\centering
\caption{Quantitative evaluation of bidirectional
registration on the FAHSYSU and PICCOLO test sets.}
\label{tab:registration_quality}
\resizebox{\columnwidth}{!}{%
\begin{tabular}{llcccc}
\toprule
Dataset & Direction
& \multicolumn{2}{c}{Cosine Similarity $\uparrow$}
& \multicolumn{2}{c}{DSC (\%) $\uparrow$} \\
\cmidrule(lr){3-4} \cmidrule(lr){5-6}
& & Before & After & Before & After \\
\midrule
FAHSYSU
& NBI$\rightarrow$WLI & 0.5433 & 0.6213 & 61.81 & 81.83 \\
& WLI$\rightarrow$NBI & 0.5433 & 0.5882 & 61.81 & 78.36 \\
\midrule
PICCOLO
& NBI$\rightarrow$WLI & 0.6200 & 0.6857 & 48.58 & 76.63 \\
& WLI$\rightarrow$NBI & 0.6200 & 0.6801 & 48.58 & 72.00 \\
\bottomrule
\end{tabular}%
}
\end{table}

To complement the quantitative evaluation, we further visualize the learned deformation fields and their effects on feature representations across multiple representative examples from both datasets. The deformation fields are learned without ground-truth displacement fields or explicit pixel-level correspondences, using segmentation supervision together with smoothness, prediction-consistency, and lesion-mask consistency constraints. Therefore, these visualizations are not intended to demonstrate physically exact anatomical motion, but to examine whether the learned transformation can establish task-relevant cross-modal correspondence for subsequent fusion.

\begin{figure}
    \centering
    \includegraphics[width=\columnwidth]{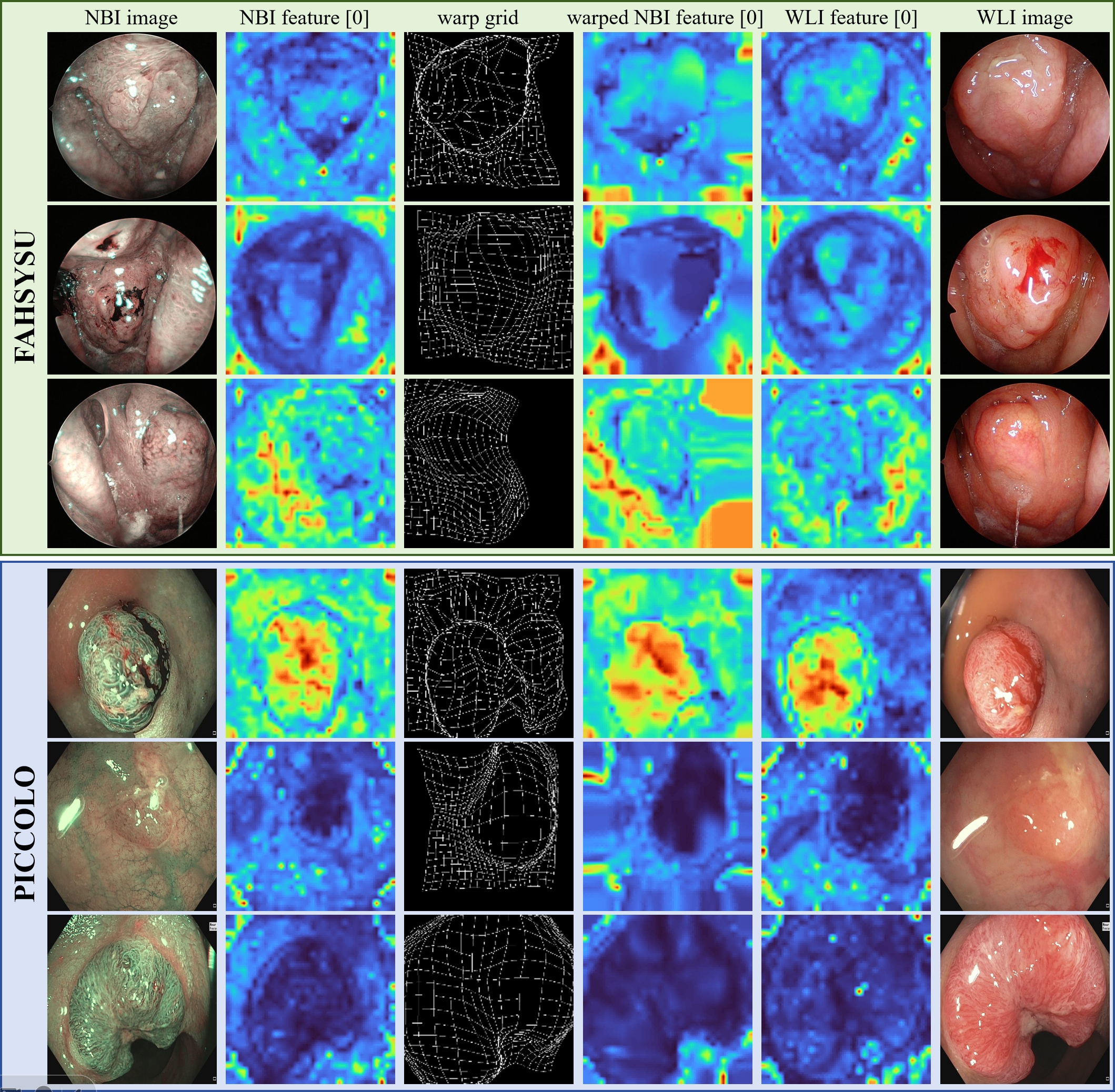}
    \caption{Feature-level registration examples from FAHSYSU and PICCOLO, illustrating learned deformation fields and improved cross-modal lesion alignment and feature correspondence after NBI-to-WLI warping.}
    \label{fig:registration1}
\vspace{-4mm}
\end{figure}

As shown in Fig.~\ref{fig:registration1}, after warping, the NBI features become more consistent with the corresponding WLI features in both lesion contour and activation distribution across the illustrated examples. This suggests that the learned deformation effectively reduces cross-modal geometric discrepancy at the representation level, consistent with the quantitative improvements reported in Table~\ref{tab:registration_quality}. Such feature-level alignment is important for our framework, because the following complex spectral fusion relies on reliable interaction between WLI-derived appearance cues and NBI-derived structural information. Overall, the quantitative and qualitative analyses support that the registration module serves as an effective intermediate mechanism for stabilizing cross-modal fusion under imperfect alignment.

\begin{figure}
    \centering
    \includegraphics[width=\columnwidth]{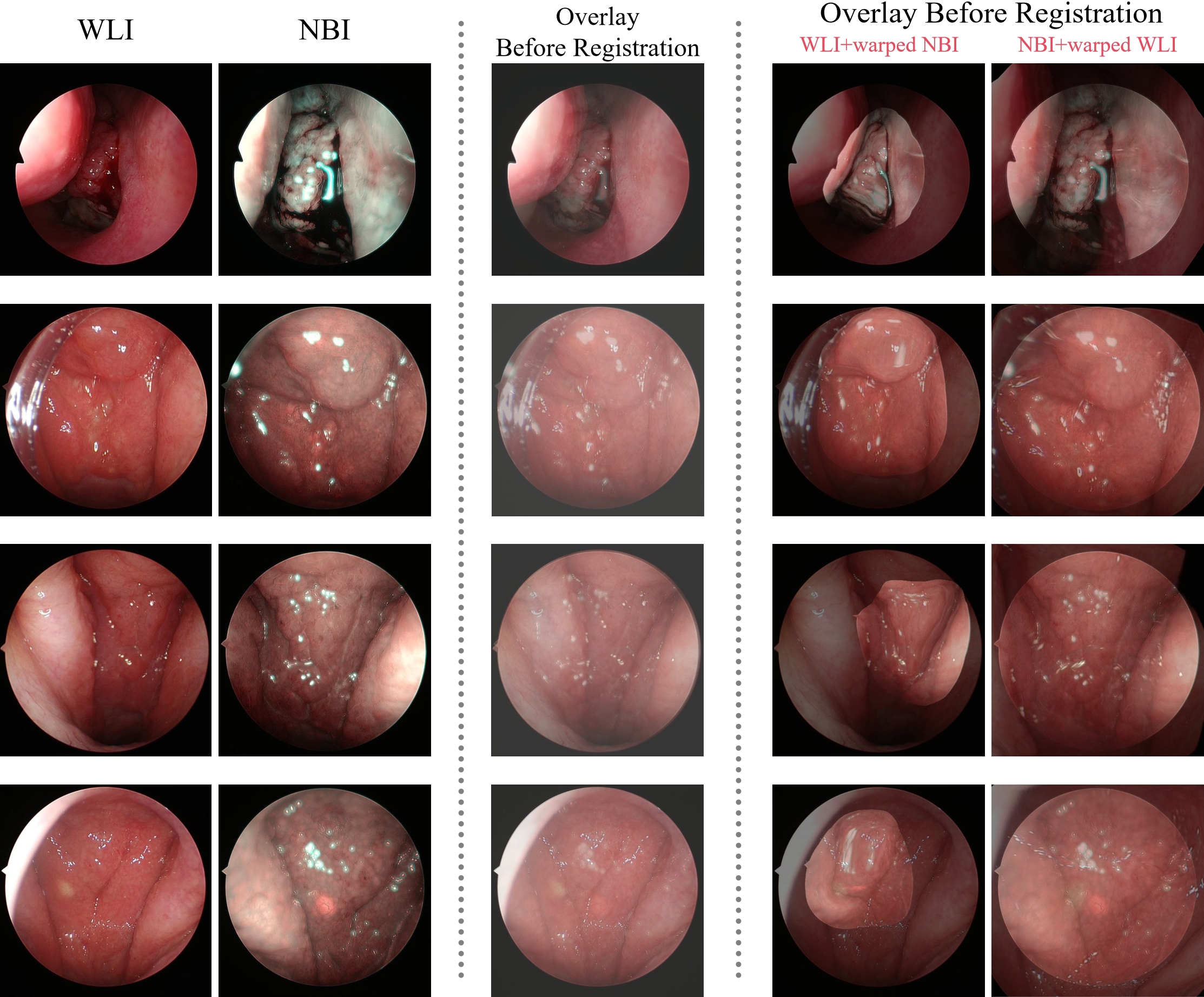}
    \caption{Representative WLI/NBI pairs with pronounced cross-modal mismatch and substantial viewpoint discrepancy, selected from the test set.}
    \label{fig:registration_before_after}
\end{figure}

To further examine registration robustness under substantial cross-modal misalignment, Fig.~\ref{fig:registration_before_after} presents four challenging WLI/NBI pairs with pronounced viewpoint differences. Compared with the pre-registration overlays, the bidirectional post-registration results exhibit improved spatial correspondence of the dominant anatomical structures and lesion regions, although some local discrepancies remain. These observations complement the test-set-level quantitative improvements in Table~\ref{tab:registration_quality}, suggesting that R-DRG can effectively alleviate pronounced cross-modal misalignment and facilitate more consistent feature interaction for subsequent multimodal fusion.

\subsection{Reliability Map and Hyperparameter Sensitivity Analysis}

\begin{figure}
\centering
\includegraphics[width=\columnwidth]{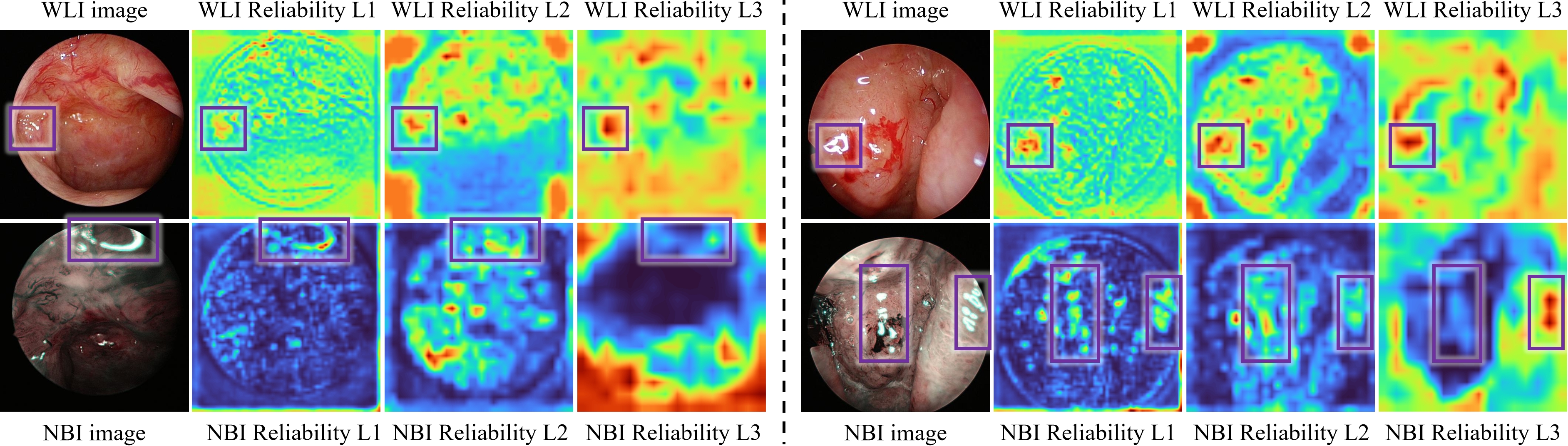}
\caption{Multi-level WLI/NBI reliability maps and their overlays on reference images. Purple boxes highlight specular reflection regions, where the learned reliability maps exhibit reduced responses, indicating sensitivity to local visual artifacts.}

\label{fig:reliability_map}
\end{figure}

To examine the spatial behavior of the proposed reliability mechanism, we visualize multi-level WLI/NBI reliability maps in Fig.~\ref{fig:reliability_map} and quantify their distributions over 3,017 FAHSYSU test pairs in Table~\ref{tab:reliability_statistics}. The 10th and 90th percentiles characterize spatial variation, while Low-Ratio measures the proportion of valid sampling locations with reliability below $e^{-1}$, indicating feature discrepancy exceeding $\tau_f$ under Eq.~(5). To assess correspondence sensitivity, we warp the counterpart lesion mask into reference coordinates using nearest-neighbor interpolation and compare reliability between mask-matched and mismatched locations within lesion-proximal regions. All statistics are averaged per pair, with the latter comparison restricted to eligible pairs containing both regions. Sampling-invalid locations are assigned zero reliability by construction and excluded from these statistics.

\begin{table}
\centering
\caption{Spatial reliability statistics on FAHSYSU (3,017 test pairs). All statistics exclude invalid sampling locations; Matched/Mismatch uses eligible pairs only.}
\label{tab:reliability_statistics}
\setlength{\tabcolsep}{4pt}
\resizebox{\columnwidth}{!}{%
\begin{tabular}{llccc}
\toprule
Branch & Level & $P_{10}/P_{90}$ & Low-Ratio (\%) & Matched/Mismatch \\
\midrule
WLI & L1 & 0.1979 / 0.7047 & 54.59 & 0.3924 / 0.3454 \\
WLI & L2 & 0.2387 / 0.8279 & 42.10 & 0.5554 / 0.5123 \\
WLI & L3 & 0.1923 / 0.5572 & 22.66 & 0.5848 / 0.5174 \\
\midrule
NBI & L1 & 0.0201 / 0.3422 & 97.51 & 0.1517 / 0.0551 \\
NBI & L2 & 0.0829 / 0.7093 & 89.54 & 0.3475 / 0.2028 \\
NBI & L3 & 0.1199 / 0.4702 & 60.03 & 0.2811 / 0.2518 \\
\bottomrule
\end{tabular}%
}
\vspace{-4mm}
\end{table}

The results reveal substantial spatial and cross-modal variation, confirming that reliability estimation does not degenerate into uniform weighting. The higher NBI Low-Ratios indicate more conservative cross-modal information transfer, rather than inherently poorer NBI quality; importantly, low reliability represents continuous attenuation rather than binary rejection. More importantly, matched regions consistently exhibit higher reliability than mismatched regions across all six configurations, with the clearest separation at NBI L2 (0.3475 vs. 0.2028). Together with the visualizations, these findings suggest that the learned reliability weights respond to residual cross-modal mismatch and enable correspondence-aware feature modulation. Nevertheless, mask disagreement remains a proxy for geometric mismatch rather than direct proof that negative transfer is eliminated.

We further investigate the sensitivity to $\tau_f$ and $\lambda_R$ by varying one parameter while fixing the other at its default value. As shown in Table~\ref{tab:hyperparameter_sensitivity}, both datasets achieve their best mean segmentation performance at $\tau_f=0.5$ and $\lambda_R=1.0$, with relatively stable results under nearby settings. Increasing $\tau_f$ excessively weakens discrepancy-dependent attenuation, whereas overly large $\lambda_R$ can overemphasize reliability in spectral reconstruction, leading to performance degradation. Notably, setting $\lambda_R=0$ does not disable reliability in the preceding magnitude--phase interpolation. Overall, these results support the importance of balanced reliability modulation for selective cross-modal fusion.

\begin{table}
\centering
\caption{Sensitivity to $\tau_f$ and $\lambda_R$ on PICCOLO and FAHSYSU. Results are mean mIoU/DSC (\%, $n=4$).}
\label{tab:hyperparameter_sensitivity}
\setlength{\tabcolsep}{4.5pt}
\renewcommand{\arraystretch}{1.12}
\begin{tabular}{cccc}
\toprule
Parameter & Value & PICCOLO & FAHSYSU \\
\midrule
\multirow{5}{*}{$\tau_f$}
& 0.25 & 82.35 / 84.31 & 80.63 / 81.99 \\
& 0.50 & \textbf{83.54 / 84.78} & \textbf{81.08 / 82.37} \\
& 1.00 & 82.77 / 84.69 & 80.92 / 82.22 \\
& 1.50 & 81.96 / 83.87 & 80.42 / 81.35 \\
& 2.00 & 81.12 / 83.05 & 80.04 / 80.99 \\
\midrule
\multirow{5}{*}{$\lambda_R$}
& 0.0 & 81.91 / 83.94 & 80.18 / 81.60 \\
& 0.5 & 82.52 / 84.45 & 80.78 / 82.12 \\
& 1.0 & \textbf{83.54 / 84.78} & \textbf{81.08 / 82.37} \\
& 2.0 & 82.43 / 84.38 & 80.69 / 82.04 \\
& 4.0 & 81.71 / 83.77 & 80.03 / 81.46 \\
\bottomrule
\end{tabular}
\vspace{-4mm}
\end{table}

\subsection{Modality-specific Segmentation Performance}

To further investigate how each modality benefits from cross-modal fusion, we separately evaluate WLI and NBI predictions on the FAHSYSU and PICCOLO test sets.
As shown in Table~\ref{tab:modality_specific}, WLI achieves higher mIoU than NBI on both datasets, whereas NBI yields slightly higher DSC.
Compared with the single-modality baselines in Table~\ref{tab:single_modal}, incorporating complementary modality information improves WLI and NBI mIoU by 3.83\% and 2.38\% on FAHSYSU, and by 7.01\% and 4.82\% on PICCOLO, respectively.
In contrast, the corresponding DSC improvements are comparable between modalities (2.53\% vs. 2.43\% on FAHSYSU and 3.00\% vs. 2.98\% on PICCOLO).
These results suggest that WLI benefits more from multimodal information in terms of mIoU, while both modalities achieve similar gains in lesion overlap accuracy.

\begin{table}
\centering
\caption{Modality-specific segmentation performance on the FAHSYSU and PICCOLO test sets.}
\label{tab:modality_specific}
\setlength{\tabcolsep}{5pt}
\renewcommand{\arraystretch}{1.1}
\begin{tabular}{llcc}
\toprule
Dataset & Modality & mIoU (\%) $\uparrow$ & DSC (\%) $\uparrow$ \\
\midrule
FAHSYSU & Overall & 81.11 & 82.44 \\
        & WLI     & 81.63 & 82.31 \\
        & NBI     & 80.59 & 82.57 \\
\midrule
PICCOLO & Overall & 82.84 & 84.77 \\
        & WLI     & 83.81 & 84.65 \\
        & NBI     & 81.87 & 84.89 \\
\bottomrule
\end{tabular}
\vspace{2pt}
\parbox{0.95\columnwidth}{
\footnotesize
Results are obtained from a single independent
run, whereas Table~\ref{Table:PICCOLO_FAHSYSU} reports mean$\pm$std over four runs. Minor differences
in the overall scores reflect run-to-run variability.
}

\end{table}

\subsection{Computational Complexity and Efficiency Analysis}

We evaluate the computational cost of each module in terms of parameters, FLOPs, and inference latency on an NVIDIA RTX 3090 GPU, using $512\times512$ WLI/NBI image pairs with a batch size of one. Latency is averaged over 100 runs after 20 warm-up iterations. As shown in Table~\ref{tab:complexity}, the full model contains 100.545M parameters and achieves an uninstrumented inference latency of 86.184 ms. R-DRG and R-CSF account for 1.62\% and 0.92\% of the total parameters, respectively, while each contributes approximately 20\% of the instrumented runtime. The main latency sources are geometric registration in R-DRG and gating/projection operations in R-CSF. Since some operations, such as feature warping and normalization, are not fully covered by FLOPs profiling, the reported FLOPs represent partial estimates, complemented by empirical runtime measurements.

\begin{table}
\centering
\caption{Module-wise computational complexity and inference latency.}
\label{tab:complexity}
\footnotesize
\setlength{\tabcolsep}{3pt}
\renewcommand{\arraystretch}{1.08}
\resizebox{\columnwidth}{!}{%
\begin{tabular}{@{}lrrr@{}}
\toprule
Module & Params (M) & GFLOPs$^\dagger$ & Time (ms) \\
\midrule
Encoder             & 88.154  & 323.493 & 48.55 \\
\midrule
R-DRG               & 1.627   & 6.884   & 20.30 \\
\quad Registration  & 1.036   & 0.530   & 14.00 \\
\quad Reliability   & 0.591   & 6.354   & 6.30 \\
\midrule
R-CSF               & 0.922   & 10.262  & 20.17 \\
\quad Complex build & 0.263   & 2.841   & 5.84 \\
\quad Spectral ops. & 0.001   & 0.342   & 2.17 \\
\quad Gate/Proj./Res.& 0.658  & 7.079   & 12.16 \\
\midrule
Decoder             & 9.842   & 62.418  & 9.10 \\
Other               & 0.000   & --      & 2.90 \\
\midrule
\textbf{Full Model} & \textbf{100.545}
                    & \textbf{403.057}
                    & \textbf{101.02} \\
\bottomrule
\end{tabular}%
}
\vspace{2pt}
\parbox{\columnwidth}{%
\scriptsize
$^\dagger$ Partially accounted FLOPs (profiled and estimated).
Time denotes instrumented inference latency.
}
\vspace{-4mm}
\end{table}

\section{Conclusion}\label{sec_Conclusion}

This paper presented a reliability-aware complex spectral fusion framework for paired-but-unregistered WLI/NBI endoscopic lesion segmentation. The proposed method addresses the key difficulty of WLI/NBI fusion by coupling topology-regularized correspondence grounding with reliability-guided complex-domain interaction. Experiments on paired WLI/NBI datasets show that our method consistently outperforms representative single-modality and multimodal segmentation baselines. These results demonstrate that selectively exploiting trustworthy cross-modal cues and modeling the distinct roles of WLI and NBI are effective for lesion segmentation under spatial misalignment.

\section{AI Assistance}

During the preparation of this work, the authors used ChatGPT (OpenAI) to assist with language polishing, grammatical correction, consistency checking, and improving the clarity and readability of the manuscript. All methodological design, scientific hypotheses, experimental results, data analyses, statistical testing, interpretation, conclusions, citations, and references were manually designed, performed, checked, and verified by the listed authors without the involvement of any large language model.
After using this tool, the authors reviewed and edited all AI-assisted language suggestions as needed and take full responsibility for the content of the manuscript.

\bibliographystyle{cas-model2-names}
\bibliography{ref}

@article{jiepy_eClinicalMedicine,
  author    = {He, Rui and Jie, Pengyu and Hou, Weijian and Long, Yudong and Zhou, Guanqun and Wu, Shumei and Liu, Wanquan and Lei, Wenbin and Wen, Weiping and Wen, Yihui},
  title     = {Real-time artificial intelligence-assisted detection and segmentation of nasopharyngeal carcinoma using multimodal endoscopic data: a multi-center, prospective study},
  journal   = {eClinicalMedicine},
  year      = {2025},
  month     = {Mar},
  day       = {01},
  volume    = {81},
  issn      = {2589-5370}
}

@article{jiepy_AoM,
author = {Rui He and Pengyu Jie and Zhangfeng Wang and Yaqiong Lu and Huanhuan Lv and Zixuan Huang and Wendong Liu and Yongquan Wang and Wanquan Liu and Wenbin Lei and Weiping Wen and Yihui Wen},
title = {Artificial intelligence-assisted real-time nasopharyngeal cancer diagnostic model enhances rhinologist performance: a prospective multi-reader study},
journal = {Annals of Medicine},
volume = {58},
number = {1},
pages = {2620335},
year = {2026},
}

@article{jiepy_TCSVT,
  author  = {Jie, Pengyu and Liu, Wanquan and Gao, Chenqiang and Wen, Yihui and He, Rui and Wen, Weiping and Li, Pengcheng and Zhang, Jintao and Meng, Deyu},
  journal = {IEEE Transactions on Circuits and Systems for Video Technology},
  title   = {A Point-Neighborhood Learning Framework for Nasal Endoscopic Image Segmentation},
  year    = {2025},
  volume  = {},
  number  = {},
  pages   = {1-1}
}

@article{Tiwari2025,
  author  = {Tiwari, Ashutosh and Mishra, Soumya and Kuo, Tsung-Rong},
  title   = {Current AI technologies in cancer diagnostics and treatment},
  journal = {Molecular Cancer},
  year    = {2025},
  month   = {Jun},
  day     = {02},
  volume  = {24},
  number  = {1},
  pages   = {159},
  issn    = {1476-4598}
}

@article{ADFNet,
  author    = {Wu, Junhao and Li, Yun and Li, Junhao and Bian, Jingliang and Fan, Xiaomao and Lei, Wenbin and Wang, Ruxin},
  title     = {Multimodal Medical Endoscopic Image Analysis via Progressive Disentangle-aware Contrastive Learning},
  journal   = {Medical Image Analysis},
  year      = {2026},
  month     = {Feb},
  day       = {28},
  volume    = {111},
  pages     = {104012}
}

@InProceedings{Su_2025_ICCV,
    author    = {Su, Dayong and Zhang, Yafei and Li, Huafeng and Li, Jinxing and Liu, Yu},
    title     = {UniFuse: A Unified All-in-One Framework for Multi-Modal Medical Image Fusion Under Diverse Degradations and Misalignments},
    booktitle = {Proceedings of the IEEE/CVF International Conference on Computer Vision (ICCV)},
    month     = {October},
    year      = {2025},
    pages     = {14238-14247}
}

@inproceedings{hu2025holistic,
  title={Holistic White-Light Polyp Classification via Alignment-Free Dense Distillation of Auxiliary Optical Chromoendoscopy},
  author={Hu, Qiang and Wang, Qimei and Chen, Jia and Ji, Xuantao and Liu, Mei and Li, Qiang and Wang, Zhiwei},
  booktitle={International Conference on Medical Image Computing and Computer-Assisted Intervention},
  pages={251--261},
  year={2025},
  organization={Springer}
}

@article{lei2025virtual,
  title={Virtual NBI image synthesis using stable diffusion for enhanced recognition of early gastric cancer: a technical validation study},
  author={Lei, Changda and Kan, Xiuji and Ouyang, Yifan and Mei, Yutong and Guo, Yunbo and Hong, Kaicheng and Li, Junbo and Wang, Bilin and Li, Rui},
  journal={Annals of Medicine},
  volume={57},
  number={1},
  pages={2523565},
  year={2025}
}

@article{DONG2025113427,
title = {A fusion network for multi-modality medical image registration with progressive feature alignment},
journal = {Knowledge-Based Systems},
volume = {317},
pages = {113427},
year = {2025},
issn = {0950-7051},
author = {Aimei Dong and Jingyuan Xu and Long Wang},
}

@article{CHEN2026103960,
title = {Joint multi-view embedding with progressive multi-scale alignment for unaligned infrared-visible image fusion},
journal = {Information Fusion},
volume = {128},
pages = {103960},
year = {2026},
issn = {1566-2535},
author = {Yida Chen and Yafei Zhang and Huafeng Li and Zhengtao Yu and Yu Liu},
}

@article{xu2023murf,
  title={Murf: Mutually reinforcing multi-modal image registration and fusion},
  author={Xu, Han and Yuan, Jiteng and Ma, Jiayi},
  journal={IEEE transactions on pattern analysis and machine intelligence},
  volume={45},
  number={10},
  pages={12148--12166},
  year={2023},
  publisher={IEEE}
}

@inproceedings{huang2022reconet,
  title={Reconet: Recurrent correction network for fast and efficient multi-modality image fusion},
  author={Huang, Zhanbo and Liu, Jinyuan and Fan, Xin and Liu, Risheng and Zhong, Wei and Luo, Zhongxuan},
  booktitle={European conference on computer Vision},
  pages={539--555},
  year={2022},
  organization={Springer}
}

@article{CAO2025114597,
title = {Multimodal image generation and fusion through content-style hybrid disentanglement},
journal = {Knowledge-Based Systems},
volume = {330},
pages = {114597},
year = {2025},
issn = {0950-7051},
author = {Xu Cao and Huanxin Zou and Jun Li and Hao Chen and Xinyi Ying and Shitian He and Yingqian Wang and Liyuan Pan},
}

@article{wang2025rfsc,
  title={RFSC: Multimodal medical image alignment fusion diagnostic classification network based on de discriminator image translation},
  author={Wang, Yacong and Peng, Xufeng and Yang, Shuyi and Zhang, Yantian and Dai, Renxiang and Meng, Xianshuang and Wu, Haitao and Shan, Fei and Ying, Jun and Wu, Jingfang and others},
  journal={Biomedical Signal Processing and Control},
  volume={109},
  pages={107905},
  year={2025},
  publisher={Elsevier}
}

@InProceedings{chen2024weakly,
    author    = {Chen, Chen and Qi, Jiahao and Liu, Xingyue and Bin, Kangcheng and Fu, Ruigang and Hu, Xikun and Zhong, Ping},
    title     = {Weakly Misalignment-free Adaptive Feature Alignment for UAVs-based Multimodal Object Detection},
    booktitle = {Proceedings of the IEEE/CVF Conference on Computer Vision and Pattern Recognition (CVPR)},
    month     = {June},
    year      = {2024},
    pages     = {26836-26845}
}

@article{JIANG2026112723,
title = {PSG-MCANet: Multi-order cross-attention modeling for multimodal fusion based on punning semantic guidance},
journal = {Pattern Recognition},
volume = {172},
pages = {112723},
year = {2026},
issn = {0031-3203},
author = {Jinlin Jiang and Gang Hu and Guanglei Sheng and Guo Wei},
}

@article{zhang2025mgnet,
  title={MGNet: RGBT tracking via cross-modality cross-region mutual guidance},
  author={Zhang, Jianming and Yang, Jing and Qin, Yu and Xiao, Zhu and Wang, Jin},
  journal={Neural Networks},
  volume={190},
  pages={107707},
  year={2025},
  publisher={Elsevier}
}

@article{chen2025evaluation,
  title={Evaluation of band selection for spectrum-aided visual enhancer (SAVE) for esophageal cancer detection},
  author={Chen, Yen-Chun and Karmakar, Riya and Mukundan, Arvind and Huang, Chien-Wei and Weng, Wei-Chun and Wang, Hsiang-Chen},
  journal={Journal of Cancer},
  volume={16},
  number={2},
  pages={470},
  year={2025}
}

@ARTICLE{10648934,
  author={Chen, Linwei and Fu, Ying and Gu, Lin and Yan, Chenggang and Harada, Tatsuya and Huang, Gao},
  journal={IEEE Transactions on Pattern Analysis and Machine Intelligence}, 
  title={Frequency-Aware Feature Fusion for Dense Image Prediction}, 
  year={2024},
  volume={46},
  number={12},
  pages={10763-10780}}

@article{yan2024decoupling,
  title={Decoupling semantic and localization for semantic segmentation via magnitude-aware and phase-sensitive learning},
  author={Yan, Qingqing and Li, Shu and He, Zongtao and Zhou, Xun and Hu, Mengxian and Liu, Chengju and Chen, Qijun},
  journal={Information Fusion},
  volume={107},
  pages={102314},
  year={2024},
  publisher={Elsevier}
}

@inproceedings{liu2025frequency,
  title={Frequency Domain-Based Diffusion Model for Unpaired Image Dehazing},
  author={Liu, Chengxu and Qi, Lu and Pan, Jinshan and Qian, Xueming and Yang, Ming-Hsuan},
  booktitle={Proceedings of the IEEE/CVF International Conference on Computer Vision},
  pages={7538--7547},
  year={2025}
}

@article{UMFSegNet,
title = {Unbiased multimodal fusion for medical image segmentation based on dual-Stream adapter},
journal = {Knowledge-Based Systems},
volume = {330},
pages = {114653},
year = {2025},
issn = {0950-7051},
author = {Haotian Lu and Mingyang Yu and Xinjian Wei and Xiaoxuan Xu and Jing Xu},
}

@inproceedings{DFormer,
title={{DF}ormer: Rethinking {RGBD} Representation Learning for Semantic Segmentation},
author={Bowen Yin and Xuying Zhang and Zhong-Yu Li and Li Liu and Ming-Ming Cheng and Qibin Hou},
booktitle={The Twelfth International Conference on Learning Representations},
year={2024}
}

@ARTICLE{FTransUNet,
  author={Ma, Xianping and Zhang, Xiaokang and Pun, Man-On and Liu, Ming},
  journal={IEEE Transactions on Geoscience and Remote Sensing}, 
  title={A Multilevel Multimodal Fusion Transformer for Remote Sensing Semantic Segmentation}, 
  year={2024},
  volume={62},
  number={},
  pages={1-15}}

@article{DCRS,
  title={Deep coupled registration and segmentation of multimodal whole-brain images},
  author={Han, Tingting and Wu, Jun and Sheng, Pengpeng and Li, Yuanyuan and Tao, ZaiYang and Qu, Lei},
  journal={Bioinformatics},
  volume={40},
  number={11},
  pages={btae606},
  year={2024}
}

@article{VI-ReID,
  title={Co-segmentation assisted cross-modality person re-identification},
  author={Huang, Nianchang and Xing, Baichao and Zhang, Qiang and Han, Jungong and Huang, Jin},
  journal={Information Fusion},
  volume={104},
  pages={102194},
  year={2024},
  publisher={Elsevier}
}

@inproceedings{ShapeConv,
  title={Shapeconv: Shape-aware convolutional layer for indoor rgb-d semantic segmentation},
  author={Cao, Jinming and Leng, Hanchao and Lischinski, Dani and Cohen-Or, Daniel and Tu, Changhe and Li, Yangyan},
  booktitle={Proceedings of the IEEE/CVF international conference on computer vision},
  pages={7088--7097},
  year={2021}
}

@ARTICLE{MCSNet,
  author={Lin, Qing and Tan, Weimin and Cai, Shilun and Yan, Bo and Li, Jichun and Zhong, Yunshi},
  journal={IEEE Transactions on Neural Networks and Learning Systems}, 
  title={Lesion-Decoupling-Based Segmentation With Large-Scale Colon and Esophageal Datasets for Early Cancer Diagnosis}, 
  year={2024},
  volume={35},
  number={8},
  pages={11142-11156},
}

@article{MFNet,
  title={A unified framework with multimodal fine-tuning for remote sensing semantic segmentation},
  author={Ma, Xianping and Zhang, Xiaokang and Pun, Man-On and Huang, Bo},
  journal={IEEE Transactions on Geoscience and Remote Sensing},
  year={2025},
  publisher={IEEE}
}

@article{UMSCS,
  title={Umscs: a novel unpaired multimodal image segmentation method via cross-modality generative and semi-supervised learning},
  author={Yang, Feiyang and Li, Xiongfei and Wang, Bo and Teng, Peihong and Liu, Guifeng},
  journal={International Journal of Computer Vision},
  volume={133},
  number={7},
  pages={4442--4464},
  year={2025},
  publisher={Springer}
}

@InProceedings{Fan_2024_CVPR,
    author    = {Fan, Xin and Wang, Xiaolin and Gao, Jiaxin and Wang, Jia and Luo, Zhongxuan and Liu, Risheng},
    title     = {Bi-level Learning of Task-Specific Decoders for Joint Registration and One-Shot Medical Image Segmentation},
    booktitle = {Proceedings of the IEEE/CVF Conference on Computer Vision and Pattern Recognition (CVPR)},
    month     = {June},
    year      = {2024},
    pages     = {11726-11735}
}

@article {HD95_ASSD,
      author = "Tim Fick and Jesse A. M. van Doormaal and Lazar Tosic and Renate J. van Zoest and Jene W. Meulstee and Eelco W. Hoving and Tristan P. C. van Doormaal",
      title = "Fully automatic brain tumor segmentation for 3D evaluation in augmented reality",
      journal = "Neurosurgical Focus",
      year = "2021",
      volume = "51",
      number = "2",
      pages= "E14",
}

@article{BIOU,
  author={Meng, Yanda and Zhang, Hongrun and Zhao, Yitian and Yang, Xiaoyun and Qiao, Yihong and MacCormick, Ian J. C. and Huang, Xiaowei and Zheng, Yalin},
  journal={IEEE Transactions on Medical Imaging}, 
  title={Graph-Based Region and Boundary Aggregation for Biomedical Image Segmentation}, 
  year={2022},
  volume={41},
  number={3},
  pages={690-701},
  }

@article{joshi2023r2net,
  title={R2Net: Efficient and flexible diffeomorphic image registration using Lipschitz continuous residual networks},
  author={Joshi, Ankita and Hong, Yi},
  journal={Medical Image Analysis},
  volume={89},
  pages={102917},
  year={2023},
  publisher={Elsevier}
}

@article{sun2024nir,
  title={Medical image registration via neural fields},
  author={Sun, Shanlin and Han, Kun and You, Chenyu and Tang, Hao and Kong, Deying and Naushad, Junayed and Yan, Xiangyi and Ma, Haoyu and Khosravi, Pooya and Duncan, James S and others},
  journal={Medical Image Analysis},
  volume={97},
  pages={103249},
  year={2024},
  publisher={Elsevier}
}

@article{chen2025spatialreg,
  title={Unsupervised learning of spatially varying regularization for diffeomorphic image registration},
  author={Chen, Junyu and Wei, Shuwen and Liu, Yihao and Bian, Zhangxing and He, Yufan and Carass, Aaron and Bai, Harrison and Du, Yong},
  journal={Medical image analysis},
  pages={103887},
  year={2025},
  publisher={Elsevier}
}

@article{chen2024mirsurvey,
  title={A survey on deep learning in medical image registration: New technologies, uncertainty, evaluation metrics, and beyond},
  author={Chen, Junyu and Liu, Yihao and Wei, Shuwen and Bian, Zhangxing and Subramanian, Shalini and Carass, Aaron and Prince, Jerry L and Du, Yong},
  journal={Medical Image Analysis},
  volume={100},
  pages={103385},
  year={2025},
  publisher={Elsevier}
}

@inproceedings{fogarollo2025implicit,
  title={Implicit Deformable Medical Image Registration with Learnable Kernels},
  author={Fogarollo, Stefano and Laimer, Gregor and Bale, Reto and Harders, Matthias},
  booktitle={International Conference on Medical Image Computing and Computer-Assisted Intervention},
  pages={249--259},
  year={2025},
  organization={Springer}
}

@article{khor2023anatomically,
  title={Anatomically constrained and attention-guided deep feature fusion for joint segmentation and deformable medical image registration},
  author={Khor, Hee Guan and Ning, Guochen and Sun, Yihua and Lu, Xu and Zhang, Xinran and Liao, Hongen},
  journal={Medical Image Analysis},
  volume={88},
  pages={102811},
  year={2023},
  publisher={Elsevier}
}

@article{PICCOLO,
  title     = {Piccolo white-light and narrow-band imaging colonoscopic dataset: A performance comparative of models and datasets},
  author    = {S{\'a}nchez-Peralta, Luisa F and Pagador, J Blas and Pic{\'o}n, Artzai and Calder{\'o}n, {\'A}ngel Jos{\'e} and Polo, Francisco and Andraka, Nagore and Bilbao, Roberto and Glover, Ben and Saratxaga, Cristina L and S{\'a}nchez-Margallo, Francisco M},
  journal   = {Applied Sciences},
  volume    = {10},
  number    = {23},
  pages     = {8501},
  year      = {2020}
}

@article{xiang2024lightweight,
  title   = {Lightweight colon polyp segmentation algorithm based on improved DeepLabV3+},
  author  = {Xiang, Shiyu and Wei, Lisheng and Hu, Kaifeng},
  journal = {Journal of Cancer},
  volume  = {15},
  number  = {1},
  pages   = {41},
  year    = {2024}
}

@ARTICLE{endo_tmi_1,
  author={Pang, Yan and Long, Yucheng and Chen, Zibin and Hu, Ying and Chen, Hao and Wang, Qiong},
  journal={IEEE Transactions on Medical Imaging}, 
  title={Endoscopic Adaptive Transformer for Enhanced Polyp Segmentation in Endoscopic Imaging}, 
  year={2026},
  volume={45},
  number={3},
  pages={987-999},
}

@ARTICLE{endo_tmi_2,
  author={Wang, Shuocheng and Liu, Jiaming and Zhu, Ruoxi and Huang, Chengkang and Jing, Minge and Fan, Yibo},
  journal={IEEE Transactions on Medical Imaging}, 
  title={A Dual-Generalization Low-Light Enhancement Framework for Capsule Endoscopy Image Restoration and Segmentation}, 
  year={2026},
  volume={45},
  number={5},
  pages={1747-1762},
}

@misc{BRENet2025,
      title={Learning Flow-Guided Registration for RGB-Event Semantic Segmentation}, 
      author={Zhen Yao and Xiaowen Ying and Zhiyu Zhu and Mooi Choo Chuah},
      year={2025},
      eprint={2505.01548},
      archivePrefix={arXiv},
      primaryClass={cs.CV},
      url={https://arxiv.org/abs/2505.01548}, 
}

@article{Patch2Space2026,
year = {2026},
month = {feb},
volume = {71},
number = {4},
pages = {045008},
author = {Tang, Zhenyu and Li, Shuaishuai and Ding, Chaowei and Wang, Jinda and Pan, Junjun and Zang, Jie},
title = {Patch2Space: a registration-free segmentation method for misaligned multimodal medical images},
journal = {Physics in Medicine \& Biology},
}

@INPROCEEDINGS{ACMSegNet2025,
  author={Ma, Liang and Wang, Jinwei and Liu, Qingming and Yin, Yingzheng},
  booktitle={2025 IEEE 8th International Conference on Electronic Information and Communication Technology (ICEICT)}, 
  title={ACMSegNet: An Affine Correction-Driven Cross-Modal Attention Network for Multimodal Remote Sensing Semantic Segmentation}, 
  year={2025},
  pages={407-412},
}

@article{TransMorph,
title = {TransMorph: Transformer for unsupervised medical image registration},
journal = {Medical Image Analysis},
volume = {82},
pages = {102615},
year = {2022},
author = {Junyu Chen and Eric C. Frey and Yufan He and William P. Segars and Ye Li and Yong Du},
}

\end{document}